\documentclass{article}
\usepackage[margin=1in]{geometry}

\usepackage{hyperref}
\usepackage{url}
\usepackage[utf8]{inputenc}

\usepackage{amsmath, amssymb, amsthm, amsfonts, bm}

\newtheorem{prop}{Proposition}

\newtheorem{theorem}{Theorem}

\usepackage{enumerate}
\usepackage{graphicx}
\usepackage{array}
\usepackage{tikz}
\usepackage{hyperref, doi}
\usepackage{epsf}
\usepackage{tabularx}
\usepackage{titling}

\usepackage[utf8]{inputenc} 
\usepackage[T1]{fontenc}    
\usepackage{float}

\usepackage{caption}
\usepackage{subcaption}
\usepackage{array}
\usepackage{algpseudocode,algorithm,algorithmicx}
\usepackage{makecell}

\usepackage{selectp}

\newcommand\blfootnote[1]{%
  \begingroup
  \renewcommand\thefootnote{}\footnote{#1}%
  \addtocounter{footnote}{-1}%
  \endgroup
}

\title{Increasing Depth Leads to U-Shaped Test Risk in Over-parameterized Convolutional Networks}   

\author{Eshaan Nichani \thanks{Laboratory for Information \& Decision Systems, and Institute for Data, Systems, and Society, Massachusetts Institute of Technology} $^{,~*}$ \and Adityanarayanan Radhakrishnan \footnotemark[1] $^{,~*}$ \and Caroline Uhler \footnotemark[1]$^{~}$ }

\thanksmarkseries{arabic}

\begin{document}

 \maketitle

\blfootnote{\hspace{-1.5mm} $^*$ Equal Contribution}

\begin{abstract}
Recent works have demonstrated that increasing model capacity through width in over-parameterized neural networks leads to a decrease in test risk.  For neural networks, however, model capacity can also be increased through depth, yet understanding the impact of increasing depth on test risk remains an open question.  In this work, we demonstrate that the test risk of over-parameterized convolutional networks is a U-shaped curve (i.e.\ monotonically decreasing, then increasing) with increasing depth.  We first provide empirical evidence for this phenomenon via image classification experiments using both ResNets and the convolutional neural tangent kernel (CNTK).  We then present a novel linear regression framework for characterizing the impact of depth on test risk, and show that increasing depth leads to a U-shaped test risk for the linear CNTK.  In particular, we prove that the linear CNTK corresponds to a depth-dependent linear transformation on the original space and characterize properties of this transformation.  We then analyze over-parameterized linear regression under arbitrary linear transformations and, in simplified settings, provably identify the depths which minimize each of the bias and variance terms of the test risk. 
\end{abstract}

\section{Introduction}
\label{sec: Introduction}

According to traditional statistical learning theory, over-parameterized models will overfit training data and thus generalize poorly to unseen test data \cite{ElementsofStatisticalLearning}.  This hypothesis is classically explained through the bias-variance tradeoff: as model complexity increases, so will variance, and thus more complex models will have increased test risk.  On the other hand, over-parameterized deep networks often perform well on test data even though they have enough capacity to interpolate randomly labeled training data~\cite{RethinkingGeneralization}.  

\begin{figure*}[!t]
    \centering
    \begin{subfigure}[t]{.45 \textwidth}
        \centering
        \includegraphics[scale=0.33]{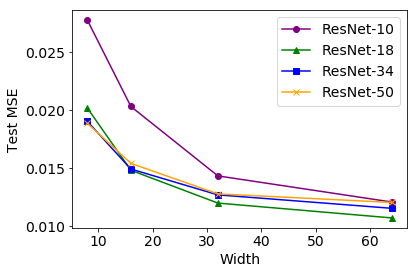}
        \caption{\centering ResNets of increasing width on CIFAR10}
    \end{subfigure}%
    \begin{subfigure}[t]{.45 \textwidth}
        \centering
        \includegraphics[scale=0.33]{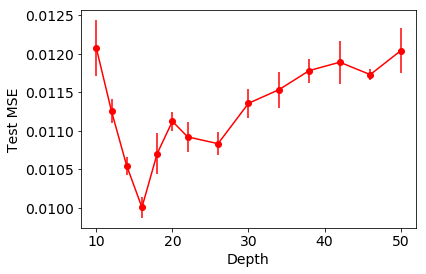}
        \caption{ ResNet of width 64 on CIFAR10.}
    \end{subfigure}%
    \caption{ (a) Increasing width in interpolating ResNets trained on CIFAR10 results in a decrease in test error. (b) In contrast, increasing the depth of ResNets trained on CIFAR10 results in a U-shaped curve (results are averaged across 3  random seeds). All models achieve 100\% training accuracy and hence are in the interpolating regime.}
    \label{fig: ResNet paper test accuracies}
\end{figure*}

Recent work \cite{DoubleDescent} reconciled this conflict via the double descent risk curve, which proposes that test risk decreases as model complexity increases beyond the interpolation threshold. While the precise behavior near the interpolation threshold depends on a number of factors including regularization~\cite{optimalregularization2020} and label noise,  several works \cite{TwoModelsDoubleDescent, SurprisesHighDimensional, BenignOverfitting, RethinkBiasVariance} proved a decrease in test risk in linear and random feature regression, when increasing the number of features used for regression.  Moreover, \cite{RethinkBiasVariance, DeepDoubleDescent, DoubleDescent} demonstrated that increasing width beyond the interpolation threshold while holding depth constant in over-parameterized neural networks leads to a decrease in test error.  This is illustrated in Figure~\ref{fig: ResNet paper test accuracies}(a), showing instances of test loss decreasing as network width increases.

While prior work analyzes increasing model complexity through width, complexity can also be increased through depth~\cite{TelgarskyDepth}. Understanding the impact of increasing depth on test error has been an open question; while linear regression provides a simple setting for understanding the behavior of test risk as a function of width, so far there does not exist a similar, simple framework for analyzing test risk through increasing depth.  

In this work, we show both empirically via CIFAR10 classification experiments and theoretically via a novel linear regression framework that the test error of over-parameterized convolutional neural networks (CNNs) follows a U-shaped curve, i.e. a curve that is monotonically decreasing until a critical threshold and then increasing.  This is illustrated in Figure \ref{fig: ResNet paper test accuracies}(b), which shows that increasing depth while holding width constant leads to a U-shaped test error in over-parameterized ResNets~\cite{ResNet}.  

The specific contributions of our paper are as follows:

\begin{enumerate}
    \item We provide empirical evidence that test error follows a U-shaped curve for interpolating CNNs (Section \ref{sec: Empirical Evidence}). We demonstrate this phenomenon for image classification on CIFAR10 \cite{CIFAR10} using over-parameterized  ResNets and infinite width convolutional networks, i.e. the convolutional neural tangent kernel (CNTK)~\cite{AroraCNTK}.  
    \item We propose a novel linear regression framework based on the linear CNTK in order to model the effect of depth on test risk (Section \ref{sec: linear neural networks}). We first derive the feature map of the linear CNTK as a function of depth.  Namely, we show that the CNTK at depth $D$ has the form $K(x, x') = x^T \Theta_D x'$ for a positive semi-definite feature transformation $\Theta_D$. We then provide a recursion to compute $\Theta_D$, and characterize properties of $\Theta_D$ for different network architectures.
    \item  In order to characterize the impact of depth on test risk for the linear CNTK, we derive the bias-variance decomposition for over-parameterized linear regression using kernels of the form $K(x, x') = x^T\Theta x'$.  We prove that the variance is minimized when $\Theta$ equals the inverse of the data covariance matrix and that the bias decreases as $\Theta$ aligns more with the true predictor.  We apply our theory to classifying MNIST digits with the linear CNTK and observe that, as predicted by our theroetical results, the test loss follows a U-shaped curve and is minimized approximately at the depth at which $\Theta_D$ is closest to the prediction direction. 
\end{enumerate}




\section{Related Work}
\label{sec: Related Work}

\noindent \textbf{Role of Width in Neural Networks.} Previous works studying generalization for over-parameterized neural networks have primarily focused on increasing width; in particular, \cite{DoubleDescent, DeepDoubleDescent}~demonstrated that double descent occurs when increasing the width of neural networks trained on MNIST \cite{mnist-lecun1998} and CIFAR10, respectively. In addition, several theoretical works~\cite{BenignOverfitting, TwoModelsDoubleDescent, OldProblemLinearRegression,  SurprisesHighDimensional, MitraLinearRegression, HarmlessInterpolation} demonstrated double descent by analyzing linear or shallow non-linear models with an increasing number of features. Moreover, \cite{RethinkBiasVariance} analyzed the bias-variance decomposition as a function of width, and showed that bias is decreasing while variance is hill shaped. In Section \ref{sec: Empirical Evidence}, we conduct a similar empirical analysis to \cite{DeepDoubleDescent, RethinkBiasVariance}, but on the impact of depth in CNNs rather than width. Our theoretical analysis in Section \ref{sec: linear neural networks}  derives the bias-variance decomposition under an arbitrary linear feature map.\\

\noindent \textbf{Role of Depth in CNNs.} Previous analyses of model scaling for ResNet~\cite{ResNet} and EfficientNet~\cite{EfficientNet} have demonstrated decreasing test loss with increasing depth. However, these experiments were not conducted in the interpolating regime, which is the focus of this paper.  We show that for CNNs which interpolate the data, test loss increases beyond a certain depth threshold.

Other works have studied the role of depth in CNNs through frameworks including expressivity~\cite{NguyenExpressivity}, sparsity~\cite{NeyshaburConvolutions}, student-teacher training~ \cite{UrbanDeepConvolutional}, and implicit regularization. Works on implicit regularization have characterized the inductive bias of over-parameterized autoencoders~\cite{RadhaAutoencoders, IdentityCrisis}, analyzed the inductive bias in function space of linear convolutional networks \cite{GunasekarLinearConv, Gunasekar2021LinearConvMulti}, and demonstrated that depth can lead to low-rank bias~\cite{Isola2021SimplicitBias}. While each of these works identified an impact of depth in CNNs, they do not provide an explicit connection to generalization, which is the focus of our work.\\

\noindent \textbf{Infinite-Width Neural Networks.} Our linear regression analysis in Section 4 is conducted in the infinite width regime. With proper scaling as width approaches infinity, the solution of training neural networks is given by the solution to a corresponding kernel regression problem, using the so-called \emph{neural tangent kernel}, or NTK \cite{JacotNTK}. In \cite{AroraCNTK}, the NTK was computed for convolutional networks (CNTK). The authors in~\cite{XiaoDynamicalIsometry, XiaoGeneralization} showed that the CNN signal propagation and CNTK (under the architectural constraints of flattening and circular padding) converge to those of a fully connected network in the limit of infinite depth. Furthermore, they provided initial empirical evidence of degradation of test risk with increasing depth. While this implies that the performance of such a CNTK must eventually degrade, in this paper we aim to understand the exact trend of depth versus test performance under different architectural constraints. We are specifically interested in identifying the optimal depth for generalization, which we accomplish in the linear CNTK setting.

\begin{figure*}[!t]
    \centering
    \includegraphics[width=\textwidth]{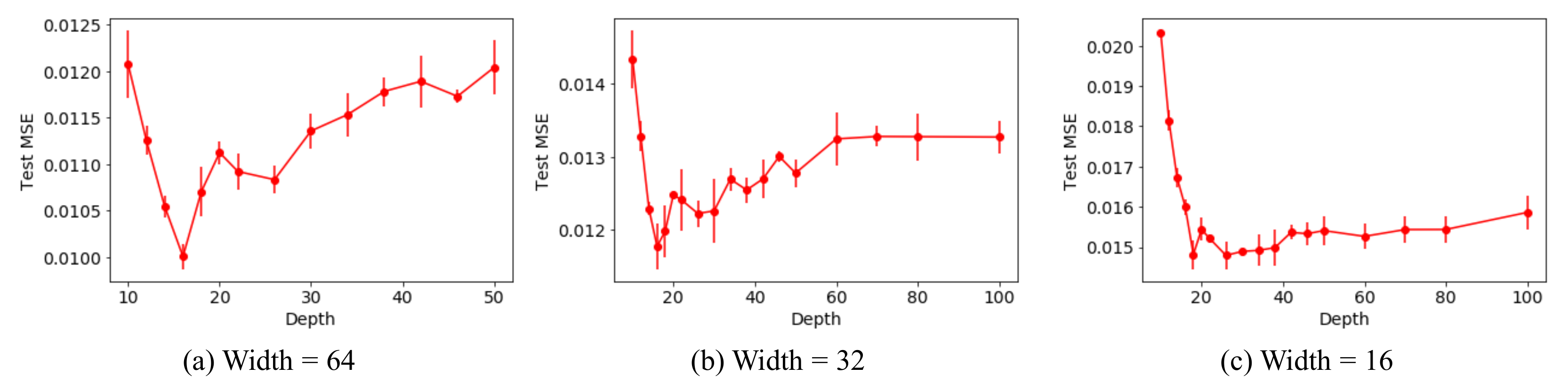}
    \caption{Test loss for ResNets of width 64 (a), width 32 (b), and width 16 (c) shows a U-shaped curve.}
    \label{fig: ResNet CIFAR10}
\end{figure*}

\section{Empirical Evidence in Nonlinear CNNs}
\label{sec: Empirical Evidence}

We begin by empirically showing that the test risk of interpolating CNNs follows a U-shaped curve when depth increases.  We illustrate this phenomenon for ResNets applied to CIFAR10 and the CNTK applied to subsets of CIFAR10.  Our training methodology is outlined in Appendix \ref{app: experiments}. In Appendix~\ref{app: fully conv}, we also demonstrate the phenomenon for finite width fully-convolutional networks applied to CIFAR10 and ImageNet-32 \cite{ImageNet32}.

\subsection{Image Classification with Modern Deep Learning Models}

To understand the effect of increasing depth in modern machine learning models, we analyzed variants of ResNet trained on CIFAR10. We consider modifications of the standard ResNet-18/ResNet-34 architectures to allow for models of arbitrary depth (see Figure \ref{fig: custom ResNet} of the Appendix for a diagram). ResNet models consist of 4 stages, each of which are connected by a downsampling operation. Each stage is comprised of a number of \textit{basic blocks}, which are two layers with a residual connection. There is a convolutional layer before the first stage, and a fully connected layer after the last stage. By varying the number of blocks in each stage, we construct ResNet models of different depths.  In particular, choosing $(n_1, n_2, n_3, n_4)$ blocks in each stage respectively produces a ResNet of depth $2 + 2\cdot(n_1 + n_2 + n_3 + n_4)$. The width $w$ of a model is defined to be the number of channels in the first stage and there are then $(w, 2w, 4w, 8w)$ channels in each stage respectively. All of our ResNet models also use batch normalization. We trained models up to depth 100, see Appendix~\ref{app: experiments} for the stage breakdown of all models used.

Our training methodology is based on that of \cite{RethinkBiasVariance}. The ResNets are trained using MSE loss on a random subset of 25,000 training images from CIFAR10. We trained for 500 epochs using SGD with learning rate $0.1$ and momentum $0.9$, and we decreased the learning rate by a factor of $10$ every 200 epochs. We also used the data augmentation scheme of random crops and random horizontal flips on the training data. For additional architectural choices and optimization methods in the fully-convolutional setting, see Appendix~\ref{app: fully conv}.

Plots of test loss as a function of depth are shown in Figure~\ref{fig: ResNet paper test accuracies} and Figure~\ref{fig: ResNet CIFAR10}. In all cases, we observe that the test loss begins to increase after a critical depth. This provides evidence for a general U-shaped dependence of test risk on depth. We emphasize that we had no optimization difficulties with training these models, and all models were able to interpolate the data (achieve near zero training loss). These experiments are therefore indeed in the interpolating regime, and hence the degradation in performance is due to the capacity of the deep models, not optimization difficulties. In Appendix~\ref{app: Additional Experiments} we provide train and test losses and accuracies of all models, and also demonstrate that other naive forms of overparameterization (increasing filter size or increasing depth in later blocks), lead to the same phenomenon, namely an increase in test loss.\\

\noindent \textbf{Remarks.} The fact that test loss can  increase with sufficient depth was present in the original ResNet paper~\cite[Table 6]{ResNet}, which demonstrated that a depth 1000 ResNet generalized worse than shallower ResNets.  On the other hand, recent works on scaling in neural networks, such as EfficientNet~\cite{EfficientNet}, have shown improving test loss with increasing depth. However, all these experiments do not achieve near zero train loss (and hence are not in the interpolating regime).  Our experiments show that for CNNs which interpolate the training data, increasing depth leads to worsening test performance.

\begin{figure*}[!t]
    \centering
    \includegraphics[width=0.8\textwidth]{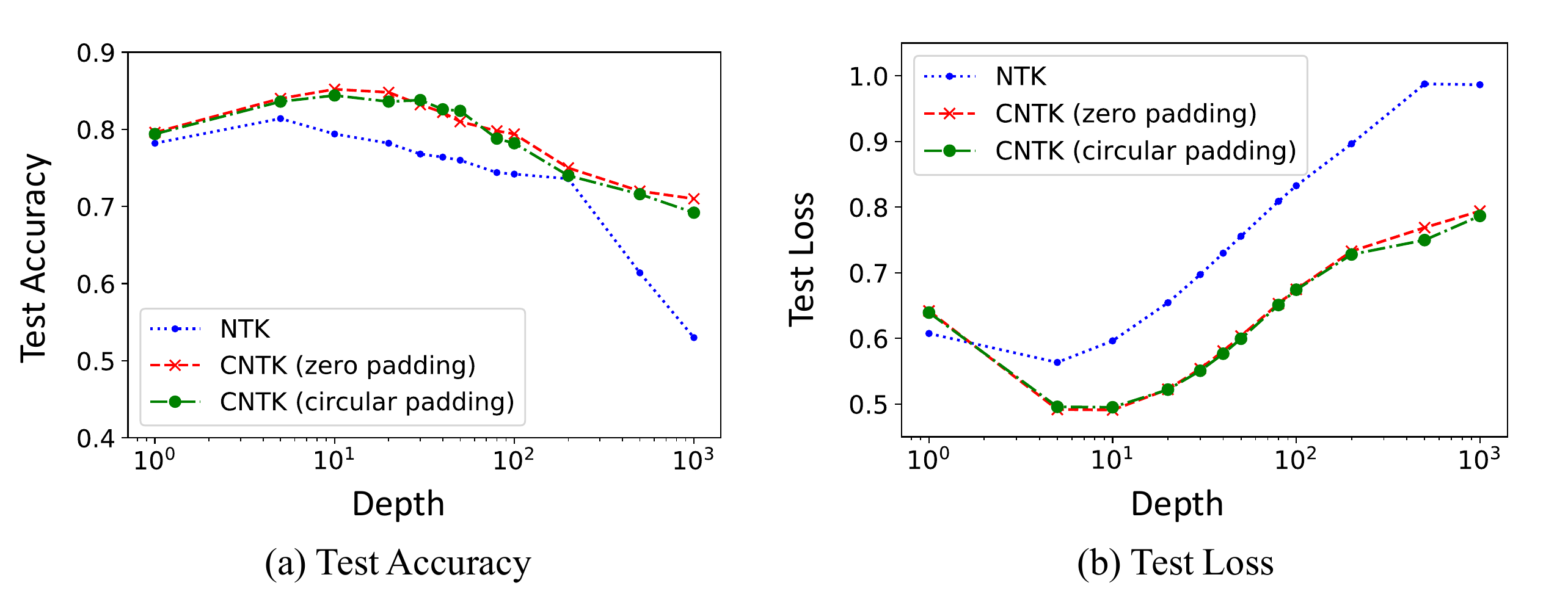}
         \caption{Test loss and accuracy for the CNTK of increasing depth (on a log scale), trained on a subset of planes and trucks from CIFAR10. Beyond a critical depth, the test error for the CNTK increases and the accuracy decreases.}
    \label{fig: CNTK}
\end{figure*}

\subsection{Image Classification with the Convolutional Neural Tangent Kernel}
\label{sec: CNTK}

We next empirically analyze the effect of depth on generalization of the CNTK, which allows us to remove the impact of additional factors in modern neural networks such as random initialization, width, batch normalization, and down-sampling. Since we are training via kernel ridge-less regression, our predictor perfectly interpolates the training data at all depths.  Our training set consists of 250 examples each of CIFAR10 planes and trucks\footnote{Computing the CNTK at large depths with more training examples was computationally prohibitive.}. We use the Neural Tangents~\cite{neuraltangents2020} library to compute the CNTK, and consider convolutional networks with filters of size $3$, both zero and circular padding, and a final flattening layer. We calculate the test loss and accuracy on a test set of 250 planes and trucks.  

In Figure~\ref{fig: CNTK}, we plot the test loss as a function of depth for both CNTK architectures, as well as the NTK. We observe that test error is monotonically decreasing up to a critical depth, after which it is monotonically increasing. This provides further evidence that such a phenomenon occurs more generally in over-parameterized convolutional networks. In Appendix~\ref{sec: CNTK extra}, we present additional empirical evidence that generalization of the CNTK worsens past a critical depth by considering different architectural choices (pooling versus flattening layers) and classification problems with a varying number of classes.\\

\noindent \textbf{Remarks.} \cite{XiaoGeneralization} proves that the CNTK with flattening and circular padding converges to the NTK as depth goes to infinity. We also conduct experiments with zero padding, which is more commonly used in practice, and provide the precise dependence of test loss on depth. This distinction is important because, as is discussed in the following section, linear CNTKs with zero padding versus circular padding have different behaviors.

\section{The Role of Depth in Linear Neural Networks}
\label{sec: linear neural networks}

In the previous section, we empirically demonstrated that CNNs have a U-shaped dependence of test risk on depth. In this section, we present a framework for understanding this phenomenon by analyzing the linear CNTK. We begin by deriving the feature map of the linear CNTK as a function of depth.  In order to characterize the test risk of the feature map for the linear CNTK, we study the bias-variance decomposition for arbitrary linear transformations, and identify the transformation that minimizes each of the bias and variance. Finally, we demonstrate how a varying transformation as a function of depth leads to a U-shaped test risk.  All proofs are presented in Appendix~\ref{app: proofs}.

\subsection{Linear CNTK Analysis}
We begin by formally defining a linear multi-channel convolutional network. We assume that the inputs have only a single channel\footnote{The case of multiple color channels can be handled by summing the CNTKs for each color channel.}, and represent them as vectors $x \in \mathbb{R}^p$. Let our network have depth $D$, and let the $d$\textsuperscript{th} layer have $C_d$ channels. For $1 \le d \le D$ and $1 \le c \le C_d$, the value of the $c$\textsuperscript{th} channel in the $d$\textsuperscript{th} layer is denoted by $x^{(d)}_c \in \mathbb{R}^p$ (we define $C_0 = 1$ and $x^{(0)}_1 = x$). We adopt the following general definition of a convolutional network to allow for varying architecture choices. Let $\{B_1, \dots, B_r\} \subset \mathbb{R}^{p \times p}$ be a basis for an $r$-dimensional subspace of $\mathbb{R}^{p \times p}$. Let $\theta_{ij}^d \in \mathbb{R}^r$ be the weights from the $i$th channel in layer $d-1$ to the $j$th channel in layer $d$. The values of the network at the $d$\textsuperscript{th} layer are defined recursively by
\begin{equation}
 \label{eq: CNTK def}
    x^{(d)}_j = \frac{1}{\sqrt{C_{d-1}}}\sum_{i=1}^{C_{d-1}}\sum_{k=1}^r\theta_{ij; k}^dB_kx^{(d-1)}_i.
\end{equation}
Let $\bm{\theta}$ be the vector of all $\theta$'s aggregated. We consider two different network architectures. In a \emph{flattening network}, all values in all channels in the final layer $x^{(D)}$ are aggregated into a single vector, followed by a single fully connected layer. Formally, the output of the network is given by
\begin{equation} \label{eq: flat def}
f^D_{flat}(x; \bm{\theta}) := \frac{1}{\sqrt{pC_D}}\sum_{c=1}^{C_D} w_c^Tx^{(D)}_c,
\end{equation}
where $w_c \in \mathbb{R}^p$. In a \emph{pooling network}~\cite{AroraCNTK}, the values in each channel in the final layer are averaged before a fully connected layer is applied to these average values. Formally, the output of the network is given by
\begin{equation} \label{eq: pool def}
    f^D_{pool}(x; \bm{\theta}) := \frac{1}{\sqrt{C_D}}\sum_{c=1}^{C_D}w_c\cdot \frac{1}{p}\sum_{i=1}^p x_{c, i}^{(D)}, 
\end{equation}
where $w_c \in \mathbb{R}$. All parameters ($\bm{\theta}$ and the $w_c$) are initialized as i.i.d standard normal, and our network is trained via gradient descent on $\bm{\theta}$ with sufficiently small learning rate using the square loss.\\

\noindent \textbf{The CNTK Limit.} In the limit as $C_1, \dots, C_D \rightarrow \infty$, the predictor learned by gradient descent with sufficiently small learning rate on the square loss is equivalent to that learned from kernel regression using the CNTK \cite{AroraCNTK}. The CNTK between two inputs $x_1, x_2 \in \mathbb{R}^p$ is defined as
\begin{equation}
    K^D(x_1, x_2) := \left \langle \frac{\partial f^D(x_1)}{\partial \bm{\theta}}, \frac{\partial f^D(x_2)}{\partial \bm{\theta}}\right \rangle,
\end{equation}
where $f^D$ is either of the two networks $f^D_{flat}$ or $f^D_{pool}$ defined above. Since the network is linear, the kernel is linear, and hence  $K^D(x_1, x_2) = x_1^T\Theta_Dx_2$ for some positive semi-definite matrix $\Theta_D \in \mathbb{R}^{p \times p}$, which we refer to as the \emph{feature transformation}. The following theorem tells us that $\Theta_D$ follows a particular recursion, with initial condition depending on the network architecture. 

\begin{theorem} 
\label{thm: linear CNTK operator}
Let the CNTK for a depth $D$ linear convolutional network be $K^D(x_1, x_2) = x_1^T\Theta_Dx_2$. Then $\Theta_D = c_D\mathcal{A}(\Theta_{D-1})$, $c_D$ is a constant depending on $D$ and $\mathcal{A}:\mathbb{R}^{p \times p} \rightarrow \mathbb{R}^{p \times p}$ is the linear operator defined as:
\begin{equation}
    \mathcal{A}(X) = \sum_{k=1}^r B_k^T X B_k.
\end{equation}
Furthermore the initial condition $\Theta_0$ depends on the architecture as follows: If $f$ is a flattening network, then $\Theta_0 = I_p$, and if $f$ is a pooling network, then $\Theta_0 = J_p$, the all-ones matrix.
\end{theorem}

We next analyze how $\Theta_D$ evolves as a function of $D$ for pooling versus flattening networks and zero versus circular padding, for 1-D convolutional networks with filter size 3 ($r = 3$). We derive the formula for 2-D convolutions in Appendix~\ref{app:2d conv}.\\

\noindent \textbf{Zero padding.} In a CNN with zero padding, the 3 basis vectors are $\{B_1\}_{ij} = \delta_{i = j - 1}$ (ones above the diagonal), $B_2 = I_n,$ and $ \{B_3\}_{ij} = \delta_{i = j + 1}$ (ones below the diagonal). For example, for $p = 4$, we have
    \[
    B_1 = \begin{bmatrix}
    0 & 1 & 0 & 0 \\
    0 & 0 & 1 & 0 \\
    0 & 0 & 0 & 1 \\
    0 & 0 & 0 & 0
    \end{bmatrix}~~~~B_2 = \begin{bmatrix}
    1 & 0 & 0 & 0\\
    0 & 1 & 0 & 0\\
    0 & 0 & 1 & 0\\
    0 & 0 & 0 & 1
    \end{bmatrix}~~~~B_3 = \begin{bmatrix}
    0 & 0 & 0 & 0 \\
    1 & 0 & 0 & 0 \\
    0 & 1 & 0 & 0 \\
    0 & 0 & 1 & 0
    \end{bmatrix}
    \]
    
It is a straightforward computation to see that the operator $\mathcal{A}$ from Theorem \ref{thm: linear CNTK operator} acts on matrix $X$ by $\mathcal{A}(X)_{ij} = \sum_{k \in \{-1, 0, 1\}}X_{i + k, j+ k}$, where $X \in \mathbb{R}^{p \times p}$ and $X_{0, i} = X_{i, 0} = X_{p+1, i} = X_{i, p+1} = 0$.   We observe that $\mathcal{A}$ acts independently on each diagonal (where $j-i$ is fixed). Let $y^m \in \mathbb{R}^{p - |m|}$ be the vector of entries where $j - i = m$ . Then, the action of $\mathcal{A}$ on $y^m$ is a $p - |m|$-dimensional Tridiagonal Toeplitz matrix where all nonzero entries are $1$. For example, when $p = 4$, $\mathcal{A}$ acts on the main diagonal $(m = 0)$ as
    \[
    \begin{bmatrix}
    \mathcal{A}(X)_{11} \\ \mathcal{A}(X)_{22} \\ \mathcal{A}(X)_{33} \\ \mathcal{A}(X)_{44}
    \end{bmatrix} = \begin{bmatrix}
    X_{11} + X_{22} \\ X_{11} + X_{22} + X_{33} \\ X_{22} + X_{33} + X_{44} \\ X_{33} + X_{44}
    \end{bmatrix}
    = \begin{bmatrix}
    1 & 1 & 0 & 0\\
    1 & 1 & 1 & 0\\
    0 & 1 & 1 & 1\\
    0 & 0 & 1 & 1
    \end{bmatrix}\begin{bmatrix}
    X_{11} \\ X_{22} \\ X_{33} \\ X_{44}
    \end{bmatrix}.
    \]
    The eigenvalues of such a $p - |m|$ dimensional tridiagonal Toeplitz matrix are~\cite{TridiagonalToeplitz}
    $$\left\{1 + 2\cos\left(\frac{\pi h}{p - |m|+1}\right) : h \in [p - |m|]\right\}$$ 
    and thus, the eigenvalues of $\mathcal{A}$ are $$\left\{1 + 2\cos\left(\frac{\pi h}{p - |m|+1}\right) : h \in [r], -p+1 \le m \le p-1 \right\}.$$
    Hence, $1 + 2\cos(\frac{\pi}{p+1})$ is the maximum eigenvalue of $\mathcal{A}$, corresponding to $m = 0, h = 1$. The corresponding eigenvector is only nonzero on the diagonal and is given by the matrix $$\Theta^* = \text{diag}\left(\sin\left(\frac{\pi}{p+1}\right), \sin\left(\frac{2\pi}{p+1}\right) \dots, \sin\left(\frac{n\pi}{p+1}\right)\right) ~, $$
    which follows from the formula for eigenvectors of a tridiagonal Toeplitz matrix from \cite{TridiagonalToeplitz}. Neither of the initial conditions for pooling $(\Theta_0 = J_p$) nor flattening $(\Theta_0 = I_p$) networks are orthogonal to $\Theta^*$. Therefore, in the limit $D \rightarrow \infty$ the CNTK for both pooling and flattening will converge (after normalizing) to the kernel $K^\infty(x_1, x_2) = x_1^T \Theta^* x_2$.  In Appendix~\ref{app:2d conv}, we show that the limiting $\Theta^*$ for 2-D convolutional networks with zero padding is also a diagonal matrix.\\

\noindent \textbf{Circular padding}. \cite{XiaoDynamicalIsometry, XiaoGeneralization} analyze networks with circular padding. In this case, the operator $\mathcal{A}$ acts as $\mathcal{A}(X)_{ij} = \sum_{k \in \{-1, 0, 1\}}X_{i + k \hspace{-1mm} \mod p, j+ k \hspace{-1mm} \mod p}$. The initial conditions for pooling and flattening are both eigenvectors of the map $\mathcal{A}$, and thus the kernel does not change with depth. The CNTK with circular padding and pooling remains a degenerate kernel ($\Theta_D = J_p$ at all depths). The CNTK with circular padding and flattening is just $K(x_1, x_2) = x_1^Tx_2$, which is the standard Euclidean kernel given by the linear NTK. This is consistent with the infinite depth analysis from \cite{XiaoGeneralization}, which showed that with circular padding, the CNTK with flattening and NTK converge to the same infinite-depth limit, while the CNTK with pooling converges to a different limit.\\

\begin{figure*}[!t]
    \centering
    \includegraphics[width=\textwidth]{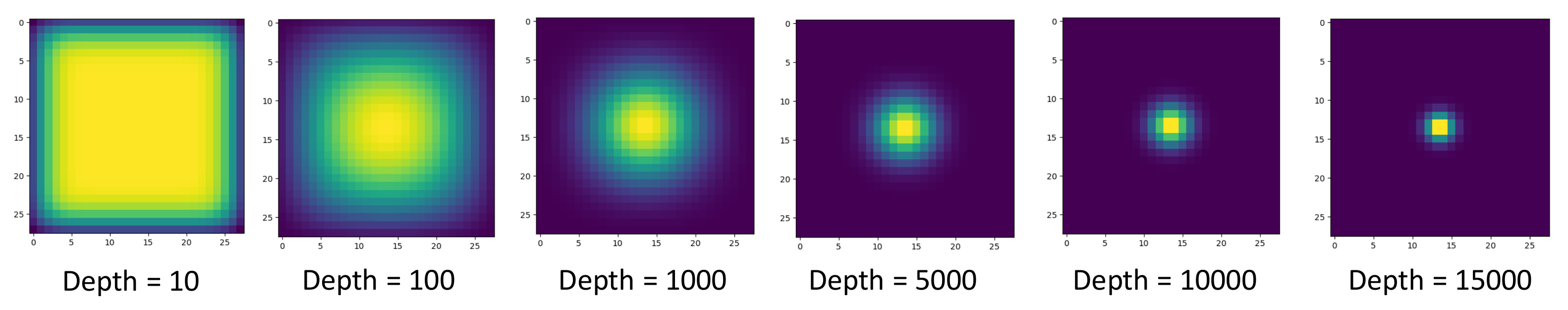}
    \caption{The leading eigenvector of $\Theta^D$ as a function of $D$ for 2-dimensional convolutions. As $D$ increases, the leading direction aggregates fewer nearby pixels, and at infinite depth, $\Theta^D$ is a diagonal matrix that simply re-weights pixels.}
    \label{fig: leading directions}
\end{figure*}

\noindent \textbf{Remarks. } We now discuss how the form of the feature transformation provides intuition for the U-shaped dependence of test loss on depth. In the case of zero padding, the infinite depth kernel $\Theta^*$ is a diagonal matrix.  Thus, this kernel simply re-scales pixels and ignores correlations between pixels, which we would expect to be sub-optimal for image classification tasks.  On the other hand, the depth zero pooling network has a degenerate kernel, which averages all the input values.  We would thus expect there to be an intermediate depth for which the kernel accurately aggregates nearby pixels at the correct scale and yields the optimal test loss.

This is demonstrated in Figure~\ref{fig: leading directions}, where we visualize the leading eigenvector of $\Theta_D$ for pooling networks on 2-D image inputs of size $28 \times 28$ ($p = 28^2 = 784, \Theta_D \in \mathbb{R}^{p \times p}$).  For increasing values of $D$, we plot the leading eigenvector of $\Theta_D$ as a $28 \times 28$ image.  For small depths, the leading direction in $\Theta_D$ averages across many pixels.  For large depths the leading direction of $\Theta_D$ is just a single pixel, which is in accordance with the preceding theory.  For intermediate depths, the most significant direction aggregates pixels at a fixed length, and thus we would expect the best performing kernel to be the one with length chosen optimally. This intuition is made precise in section \ref{sec: Regression with Arbitrary Linear Feature Maps}.\\

\noindent \textbf{Remarks on the Impact of Padding.} The feature transformations for zero versus circular padding differ in that the linear CNTK with circular padding does not change with depth while the linear CNTK with zero padding does change with depth.  Hence, the linear CNTK for networks with zero padding is a simple setting in which we can analyze the impact of depth on test error.  Furthermore, the infinite-depth limit for the zero-padding CNTK with pooling and flattening are equivalent, but differ from the NTK. When using circular padding, the infinite depth kernel for flattening converges to the NTK, but differs from the kernel for pooling.  This is consistent with the result of \cite{XiaoDynamicalIsometry, XiaoGeneralization}, which establishes the convergence of CNTK to the NTK for networks with circular padding.

\subsection{Over-parameterized Regression with Arbitrary Linear Feature Maps}
\label{sec: Regression with Arbitrary Linear Feature Maps}

\begin{figure*}[!t]
    \centering
    \begin{subfigure}[t]{.45 \textwidth}
        \centering
        \includegraphics[scale=0.4]{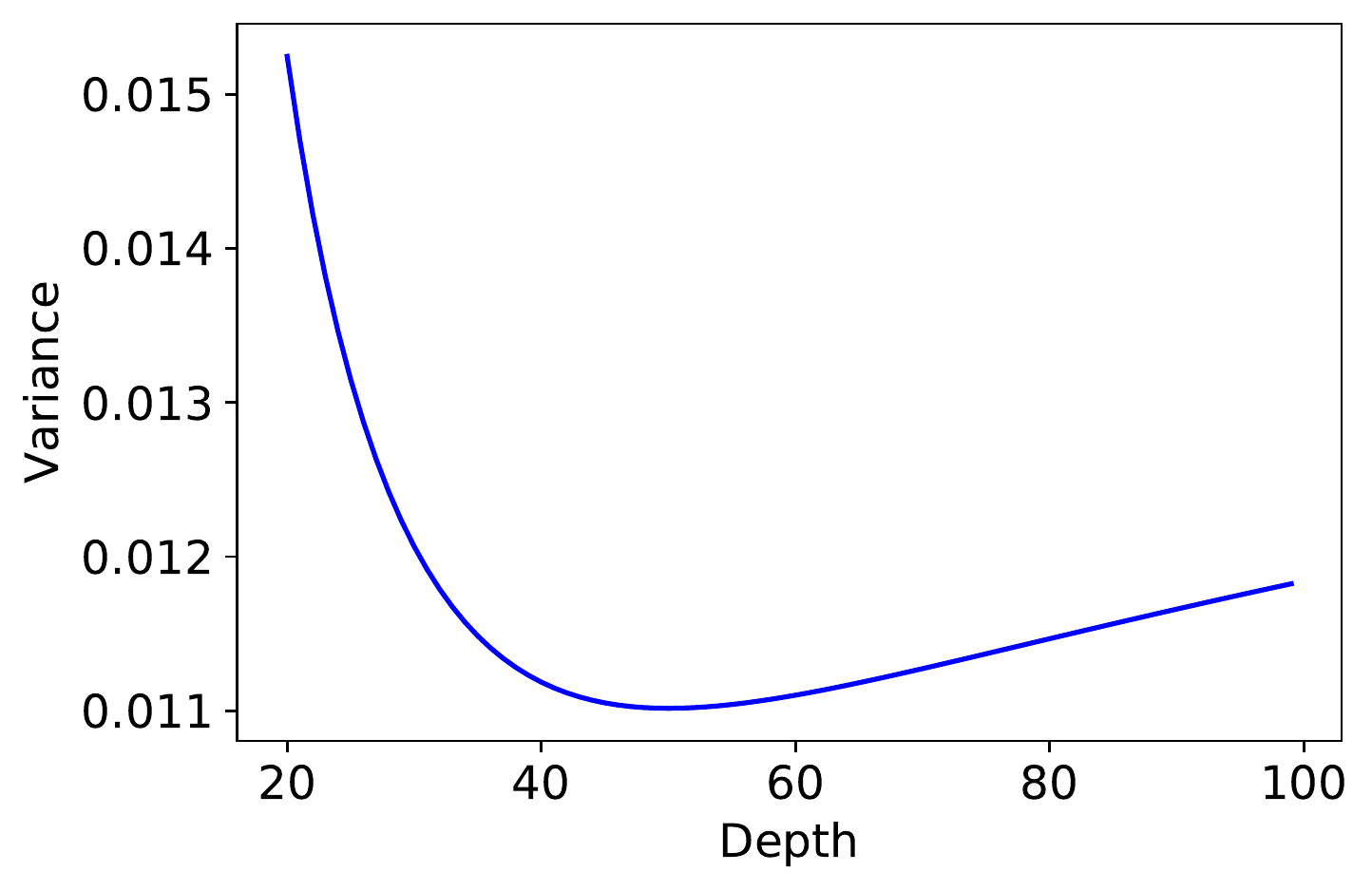}
        \caption{\centering Variance as a function of depth with $\Sigma^{-1} \propto \Theta^{50}$.}
    \end{subfigure}%
    \begin{subfigure}[t]{.45 \textwidth}
        \centering
        \includegraphics[scale=0.4]{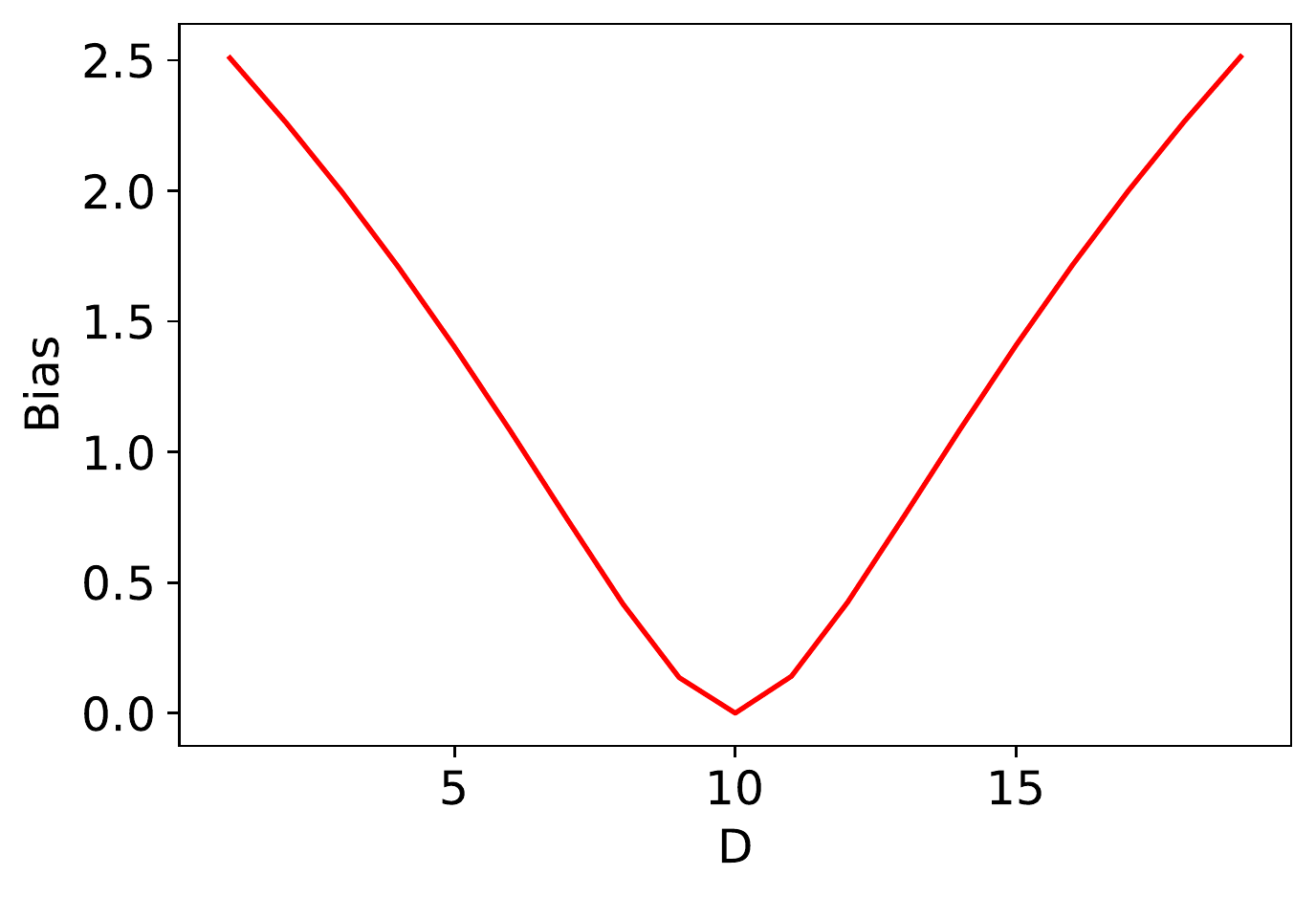}
        \caption{\centering Bias as a function of $D$ with $\Sigma = I$, $\Theta_D = \beta\beta^T + |D-10|\cdot I$.}
    \end{subfigure}%
    \caption{Over-parameterized linear regression with a linear kernel. The variance is minimized at $\Theta = \Sigma^{-1}$, and the bias is minimized as $\Theta$ approaches  $\beta\beta^T$. }
    \label{fig: bias variance theory}
\end{figure*}

In the previous section, we showed that the linear CNTK can be expressed as $K(x, x') = x^T\Theta_D x'$ for varying $\Theta_D$.  In order to connect increasing depth with test risk in the linear CNTK, we now analyze the risk of linear kernels with general feature transformation $\Theta$.  We compute the $\Theta$ which minimize each of the bias and variance, and use the form of the minimizers to demonstrate a U-shaped test risk for the linear CNTK. 

We use the following standard setup for over-parameterized linear regression.  Let  $X \in \mathbb{R}^{n \times p}$ be the data matrix and $Y \in \mathbb{R}^n$ be the response, where $p$ is the number of features, $n$ is the number of samples, and $n < p$.  Let $\Sigma \in \mathbb{R}^{p \times p}$ be the true data covariance and $\beta \in \mathbb{R}^p$ be the true predictor. Hence $x_i \overset{i.i.d}{\sim} \mathcal{N}(0, \Sigma)$, $y_i = x_i^T\beta + \varepsilon_i$, where $\varepsilon_i \overset{i.i.d.}{\sim} \mathcal{N}(0, \sigma^2)$. We thus have that $Y = X\beta + \eta$, where $\eta \sim \mathcal{N}(0, \sigma^2 I_n)$. Rather than performing ordinary least squares, we solve kernel ridge-less regression using the linear kernel $K(x, x') = x^T\Theta x'$ for arbitrary $\Theta \in \mathbb{R}^{p \times p}$, which yields the predictor
\begin{equation}
    \hat{f}(x) = x^T\Theta X^T(X\Theta X^T)^{-1}Y.
\end{equation} 
Our goal is to compute the expected excess risk of this predictor on a new sample $x \sim \mathcal{N}(0, \Sigma)$:
\begin{equation}
    \mathcal{R}(\Theta; \beta) := \mathbb{E}_{x, X, Y}\left[(x^T\beta - \hat{f}(x))^2\right].
\end{equation}
The excess risk is given by the following bias-variance decomposition (with proof in Appendix \ref{appendix: Proof of Theorem 2}): 
\begin{theorem}
The excess risk $\mathcal{R}(\Theta; \beta)$ can be written as 
\begin{equation}
    \mathcal{R}(\Theta; \beta) = \mathcal{B}(\Theta; \beta) + \mathcal{V}(\Theta).
\end{equation}
The squared bias, $\mathcal{B}(\Theta; \beta)$,  is given by
\begin{equation}
    \mathcal{B}(\Theta; \beta) = \mathbb{E}_{X}\|P_{\Theta; X}^{\perp}\beta\|_{\Sigma}^2,
\end{equation}
where $P_{\Theta; X}^{\perp} := I - \Theta X^T(X\Theta X^T)^{-1}X$ is a rotated projection operator and $\|z\|^2_\Sigma := z^T\Sigma z$.\\

\noindent The variance $\mathcal{V}(\Theta)$ is given by
\begin{equation}
    \mathcal{V}(\Theta) = \sigma^2\mathbb{E}_{z_i \sim \mathcal{N}(0, I)}\text{Tr}\left[(Z\Tilde{\Sigma}Z^T)^{-2}Z\Tilde{\Sigma}^2Z^T\right],
\end{equation}
where the rows of $Z$ are sampled i.i.d from $\mathcal{N}(0, I)$, and $\Tilde{\Sigma} = \Sigma^{1/2}\Theta\Sigma^{1/2}.$\footnote{Note that scaling $\Theta$ leaves both the bias and variance unchanged.}
\end{theorem}

We now compute the $\Theta$ which minimizes the bias and the variance respectively.  First, we show that the variance is minimized when $\Theta $ is proportional to $\Sigma^{-1}$ (proof in Appendix \ref{appendix: Proof of Prop 1}).
\begin{prop}\label{prop: variance}
The variance is lower bounded by 
\begin{equation}
    \mathcal{V}(\Theta) \ge \sigma^2 \mathbb{E}[\text{Tr}(ZZ^T)^{-1}] = \sigma^2 \frac{n}{p-n-1},
\end{equation} 
with equality if and only if $\Theta = c\Sigma^{-1}$ for a constant $c$.
\end{prop}
Next, we show that the bias decreases when $\Theta$ approaches $\beta\beta^T$ (proof in Appendix \ref{appendix: Proof of Prop 2}).
\begin{prop}\label{prop: bias}
For any positive semi-definite $\Theta$, and $0 \le \lambda \le 1$, we have that
\begin{equation}
    \mathcal{B}(\Theta; \beta) \ge \mathcal{B}((1 - \lambda)\Theta + \lambda\beta\beta^T; \beta).
\end{equation}

\end{prop}

Hence, aligning the feature transformation $\Theta$ with $\beta$ leads to a smaller bias.\footnote{We note that if $\text{rank}(\Theta) < n$ and $\beta \in \text{span}(\Theta)$, our problem is reduced to under-parameterized regression and hence the bias is zero.}  Propositions~\ref{prop: variance} and \ref{prop: bias} are verified empirically in Figure~\ref{fig: bias variance theory}. In Figure~\ref{fig: bias variance theory}(a), we let $\Theta_D$ be the feature transformation for a depth $D$ linear CNTK with 1-D inputs and $p = 20$, and use $n = 10$ training samples.  We choose the data covariance to be proportional to $\Theta_{50}^{-1}$, and let the noise variance be $\sigma^2 = 0.01$.  We observe that the variance is indeed U-shaped as a function of depth, and is minimized at depth 50.  In Figure~\ref{fig: bias variance theory}(b), we set the data covariance to be the identity, and define our family of feature transformations by $\Theta_D = \beta\beta^T + |D - 10|\cdot I_{20}$. As is proven by Proposition~\ref{prop: bias}, the bias is U-shaped as a function of depth, and is minimized at $D = 10$.

\subsection{Experiments}

\begin{figure*}[!t]
    \centering
    \includegraphics[width=0.8\textwidth]{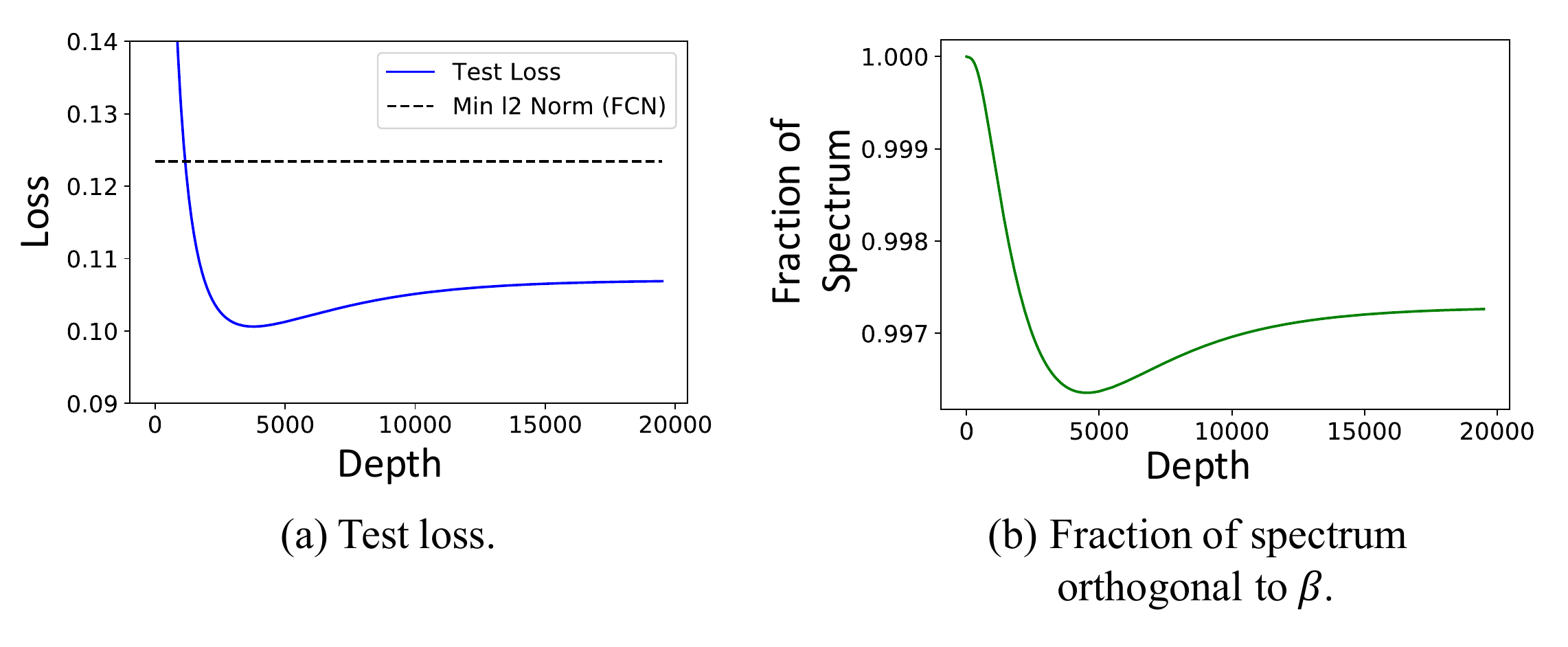}
    \caption{Linear Regression on 0/1 MNIST Digits. The test loss is U-shaped as a function of depth, with minimum test loss approximately at the depth where $\Theta_D$ points most in the direction of $\beta$. The dashed black line in (a) is the test loss of the minimum $\ell_2$ norm solution (i.e. $K(x, x') = x^Tx'$).}
    \label{fig: mnist regression}
\end{figure*}

Propositions~\ref{prop: variance} and \ref{prop: bias} imply that the bias is minimized when $\Theta_D$ aligns with $\beta$ and the variance is minimized when $\Theta_D = \Sigma^{-1}$.  In general, neither of these conditions may hold with exact equality.  Nevertheless, we now empirically show that for $\Theta_D$ given by the linear CNTK, bias is approximately minimized at the depth $D$ for which $\Theta_D$ is closest to $\beta\beta^T$.  


 
We consider linear regression on digits $0$ and $1$ from MNIST.\footnote{We use label $y = 1$ for the zeros and $y = -1$ for the ones.} To select a ground truth vector, $\beta$, we sample 50 zeros and 50 ones, and solve for $\beta$ which fits the data exactly. These 100 points thus make up the true distribution of the data. To compute the bias in this setting, we sample $n = 10$ points, fit a linear predictor using $\Theta_D$ as $D$ varies, and calculate the loss on all 100 points. The average loss (over 20 random trials) is plotted in Figure~\ref{fig: mnist regression}(a). We again observe a U-shaped dependence of test loss on depth. Proposition 2 implies that aligning $\Theta_D$ more with $\beta$ will lead to a smaller bias.  This is demonstrated in Figure~\ref{fig: mnist regression}(b), where we plot the fraction of the spectrum of $\Theta_D$ orthogonal to $\beta$. Namely, if $u = \beta/\|\beta\|$, we plot $g(D) = 1 - (u^T\Theta_Du)^2/\|\Theta_D\|_F^2$, which remains unchanged under scalar multiplication on $\Theta_D$ and ranges from 0 (if $\Theta_D$ is a multiple of $\beta\beta^T$) to 1 (if $\beta$ is in the null space of $\Theta_D$).  We observe that $g(D)$ is also U-shaped, and the depth which minimizes the bias is approximately equal to the depth which minimizes $g(D)$. To tie this back to our analysis in section 4.1, we expect that on image data the true $\beta$ vector involves aggregating nearby pixels at the correct (intermediate) length scale. Hence we expect that the $\Theta_D$ which aligns best with the true $\beta$ would be at some intermediate value of $D$. This therefore provides a mechanism through which the test risk of CNNs has a U-shaped dependence on depth.

\section{Discussion}
\label{sec: discussion}
In this work, we demonstrated that increasing neural network capacity through depth leads to a U-shaped curve for test risk.  This result is in contrast to increasing neural network capacity through width, which leads to a decrease in test risk beyond the interpolation threshold \cite{DoubleDescent}.  We first empirically showed that the test error follows a U-shaped curve in finite and infinite width nonlinear CNNs.  To theoretically analyze the effect of depth on the test risk, we presented a novel linear regression framework based on the linear CNTK.  We derived the bias-variance decomposition of using linear kernels, i.e. $K(x, x') = x^T \Theta x'$, to solve linear regression.   We then showed that the variance is minimized when $\Theta$ matches the inverse of the data covariance and the bias is minimized when $\Theta$ aligns with the optimal predictor.  We applied our theory to MNIST digit classification with the linear CNTK and observed that the test loss is approximately minimized at the depth at which the feature transformation $\Theta$ is closest to the prediction direction. 

While we identified the depth that minimizes the bias and variance when solving linear regression with linear kernels, one direction of future work is to provide a closed form for the bias and variance as a function of the feature transformation, possibly in the limit as the dimension and sample size go to infinity.  Furthermore, our theoretical analysis was in the setting of linear kernels, and thus another important direction of future work is to understand how to extend these insights to nonlinear kernels and, in particular, the nonlinear CNTK.

\section*{Acknowledgements}
The authors were supported by the National Science Foundation (DMS-1651995), Office of Naval Research (N00014-17-1-2147 and N00014-18-1-2765), MIT-IBM Watson AI Lab, and a Simons Investigator Award  to C.~Uhler. The Titan Xp used for this research was donated by the NVIDIA Corporation.

\bibliography{main}

\begin{thebibliography}{10}

\bibitem{AroraCNTK}
Sanjeev Arora, Simon~S. Du, Wei Hu, Zhiyuan Li, Ruslan Salakhutdinov, and
  Ruosong Wang.
\newblock On exact computation with an infinitely wide neural net.
\newblock In {\em Advances in Neural Information Processing Systems (NeurIPS)},
  2019.

\bibitem{BenignOverfitting}
Peter~L. Bartlett, Philip~M. Long, Gábor Lugosi, and Alexander Tsigler.
\newblock {Benign overfitting in linear regression}.
\newblock {\em Proceedings of the National Academy of Sciences}, 2020.

\bibitem{DoubleDescent}
Mikhail Belkin, Daniel Hsu, Siyuan Ma, and Soumik Mandal.
\newblock Reconciling modern machine-learning practice and the classical
  bias{\textendash}variance trade-off.
\newblock {\em Proceedings of the National Academy of Sciences},
  116(32):15849--15854, 2019.

\bibitem{TwoModelsDoubleDescent}
Mikhail Belkin, Daniel Hsu, and Ji~Xu.
\newblock {Two models of double descent for weak features}.
\newblock {\em arXiv:1903.07571}, 2019.

\bibitem{OldProblemLinearRegression}
K~Bibas, Y.~Fogel, and M.~Feder.
\newblock A new look at an old problem: A universal learning approach to linear
  regression.
\newblock {\em arXiv preprint arXiv:1905.04708}, 2019.

\bibitem{ImageNet32}
Patryk Chrabaszcz, Ilya Loshchilov, and Frank Hutter.
\newblock A downsampled variant of imagenet as an alternative to the cifar
  datasets.
\newblock {\em arXiv preprint arXiv:1707.08819}, 2017.

\bibitem{GunasekarLinearConv}
Suriya Gunasekar, Jason~D. Lee, Daniel Soudry, and Nathan Srebro.
\newblock Implicit bias of gradient descent on linear convolutional networks.
\newblock In {\em Advances in Neural Information Processing Systems (NeurIPS)},
  2018.

\bibitem{SurprisesHighDimensional}
Trevor Hastie, Andrea Montanari, Saharon Rosset, and Ryan~J Tibshirani.
\newblock Surprises in high-dimensional ridgeless least squares interpolation.
\newblock {\em arXiv preprint arXiv:1903.08560}, 2019.

\bibitem{ElementsofStatisticalLearning}
Trevor Hastie, Robert Tibshirani, and Jerome Friedman.
\newblock {\em {The Elements of Statistical Learning}}, volume~1.
\newblock Springer, 2001.

\bibitem{ResNet}
Kaiming He, Xiangyu Zhang, Shaoqing Ren, and Jian Sun.
\newblock {Deep residual learning for image recognition}.
\newblock In {\em Computer Vision and Pattern Recognition (CVPR)}, 2016.

\bibitem{Isola2021SimplicitBias}
Minyoung Huh, Hossein Mobahi, Richard Zhang, Brian Cheung, Pulkit Agrawal, and
  Phillip Isola.
\newblock The low-rank simplicity bias in deep networks.
\newblock {\em arXiv preprint arXiv:2103.10427}, 2021.

\bibitem{BatchNorm}
Sergey Ioffe and Christian Szegedy.
\newblock Batch normalization: Accelerating deep network training by reducing
  internal covariate shift.
\newblock In {\em International Conference in Machine Learning (ICML)}, 2015.

\bibitem{JacotNTK}
Arthur Jacot, Franck Gabriel, and Clément Hongler.
\newblock Neural tangent kernel: Convergence and generalization in neural
  networks.
\newblock In {\em Advances in Neural Information Processing Systems (NeurIPS)},
  2018.

\bibitem{Gunasekar2021LinearConvMulti}
Meena Jagadeesan, Ilya Razenshteyn, and Suriya Gunasekar.
\newblock Inductive bias of multi-channel linear convolutional networks with
  bounded weight norm.
\newblock {\em arXiv preprint arXiv:2102.12238}, 2021.

\bibitem{Adam}
Diederik~P. Kingma and Jimmy Ba.
\newblock {Adam: A method for stochastic optimization}.
\newblock In {\em International Conference on Learning Representations (ICLR)},
  2015.

\bibitem{CIFAR10}
Alex Krizhevsky.
\newblock {Learning multiple layers of features from tiny images}.
\newblock Master's thesis, University of Toronto, 2009.

\bibitem{mnist-lecun1998}
Yann LeCun, L{\'e}on Bottou, Yoshua Bengio, and Patrick Haffner.
\newblock {Gradient-based learning applied to document recognition}.
\newblock {\em Proceedings of the IEEE}, 86(11):2278--2324, 1998.

\bibitem{MitraLinearRegression}
Partha~P. Mitra.
\newblock Understanding overfitting peaks in generalization error: Analytical
  risk curves for $l_2$ and $l_1$ penalized interpolation.
\newblock {\em arXiv preprint arXiv:1906.03667}, 2019.

\bibitem{HarmlessInterpolation}
Vidya Muthukumar, Kailas Vodrahalli, Vignesh Subramanian, and Anant Sahai.
\newblock Harmless interpolation of noisy data in regression.
\newblock {\em IEEE Journal on Selected Areas in Information Theory},
  1(1):67--83, 2020.

\bibitem{DeepDoubleDescent}
Preetum Nakkiran, Gal Kaplun, Yamini Bansal, Tristan Yang, Boaz Barak, and Ilya
  Sutskever.
\newblock Deep double descent: Where bigger models and more data hurt.
\newblock In {\em International Conference in Learning Representations (ICLR)},
  2020.

\bibitem{optimalregularization2020}
Preetum Nakkiran, Prayaag Venkat, Sham Kakade, and Tengyu Ma.
\newblock Optimal regularization can mitigate double descent.
\newblock In {\em International Conference on Learning Representations (ICLR)},
  2021.

\bibitem{NeyshaburConvolutions}
Benham Neyshabur.
\newblock Towards learning convolutions from scratch.
\newblock In {\em Advances in Neural Information Processing Systems (NeurIPS)},
  2020.

\bibitem{NguyenExpressivity}
Quynh Nguyen and Matthias Hein.
\newblock Optimization landscape and expressivity of deep cnns.
\newblock In {\em International Conference in Machine Learning (ICML)}, 2018.

\bibitem{TridiagonalToeplitz}
Silvia Noschese, Lionello Pasquini, and Lothar Reichel.
\newblock Tridiagonal toeplitz matrices: properties and novel applications.
\newblock {\em Numerical Linear Algebra with Applications}, 20(2):302--326,
  2013.

\bibitem{neuraltangents2020}
Roman Novak, Lechao Xiao, Jiri Hron, Jaehoon Lee, Alexander~A. Alemi, Jascha
  Sohl-Dickstein, and Samuel~S. Schoenholz.
\newblock Neural tangents: Fast and easy infinite neural networks in python.
\newblock In {\em International Conference on Learning Representations (ICLR)},
  2020.

\bibitem{RadhaAutoencoders}
Adityanarayanan Radhakrishnan, Mikhail Belkin, and Caroline Uhler.
\newblock {Memorization in overparameterized autoencoders}.
\newblock In {\em ICML Workshop on Identifying and Understanding Deep Learning
  Phenomena}, 2019.

\bibitem{EfficientNet}
Mingxing Tan and Quoc~V. Le.
\newblock Efficientnet: Rethinking model scaling for convolutional neural
  networks.
\newblock In {\em International Conference on Machine Learning (ICML)}, 2019.

\bibitem{TelgarskyDepth}
Matus Telgarsky.
\newblock Benefits of depth in neural networks.
\newblock In {\em Conference on Learning Theory}, 2016.

\bibitem{UrbanDeepConvolutional}
Gregor Urban, Krzysztof~J. Geras, Samira~Ebrahimi Kahou, Ozlem Aslan, Shenjie
  Wang, Abdelrahman Mohamed, Matthai Philipose, Matt Richardson, and Rich
  Caruana.
\newblock Do deep convolutional nets really need to be deep and convolutional?
\newblock In {\em International Conference on Learning Representations (ICLR)},
  2017.

\bibitem{XiaoDynamicalIsometry}
Lechao Xiao, Yasaman Bahri, Jascha Sohl-Dickstein, Samuel Schoenholz, and
  Jeffrey Pennington.
\newblock Dynamical isometry and a mean field theory of cnns: How to train
  10,000-layer vanilla convolutional neural networks.
\newblock In {\em International Conference on Machine Learning (ICML)}, 2018.

\bibitem{XiaoGeneralization}
Lechao Xiao, Jeffrey Pennington, and Samuel Schoenholz.
\newblock Disentangling trainability and generalization in deep neural
  networks.
\newblock In {\em International Conference on Machine Learning (ICML)}, 2020.

\bibitem{LeakyReLU}
Bing Xu, Naiyan Wang, Tianqi Chen, and Mu~Li.
\newblock {Empirical Evaluation of Rectified Activations in Convolution
  Network}, 2015.
\newblock arXiv:1505.00853.

\bibitem{RethinkBiasVariance}
Zitong Yang, Yaodong Yu, Chong You, Jacob Steinhardt, and Yi~Ma.
\newblock {Rethinking Bias-Variance Trade-off for Generalization of Neural
  Networks}.
\newblock In {\em International Conference in Machine Learning (ICML)}, 2020.

\bibitem{RethinkingGeneralization}
Chiyuan Zhang, Samy Bengio, Moritz Hardt, Benjamin Recht, and Oriol Vinyals.
\newblock {Understanding deep learning requires rethinking generalization}.
\newblock In {\em International Conference on Learning Representations (ICLR)},
  2017.

\bibitem{IdentityCrisis}
Chiyuan Zhang, Samy Bengio, Moritz Hardt, and Yoram Singer.
\newblock {Identity crisis: Memorization and generalization under extreme
  overparameterization}.
\newblock In {\em International Conference on Learning Representations (ICLR)},
  2020.

\end{thebibliography}
\bibliographystyle{plain}

\clearpage
\appendix

\section{Experimental Details}
\label{app: experiments}
All models were trained on an NVIDIA TITAN RTX GPU using the PyTorch library. A repository with code used for this paper can be found here: \url{github.com/eshnich/deep_cnn}.
\begin{figure*}[ht]
    \centering
    \includegraphics[width=\textwidth]{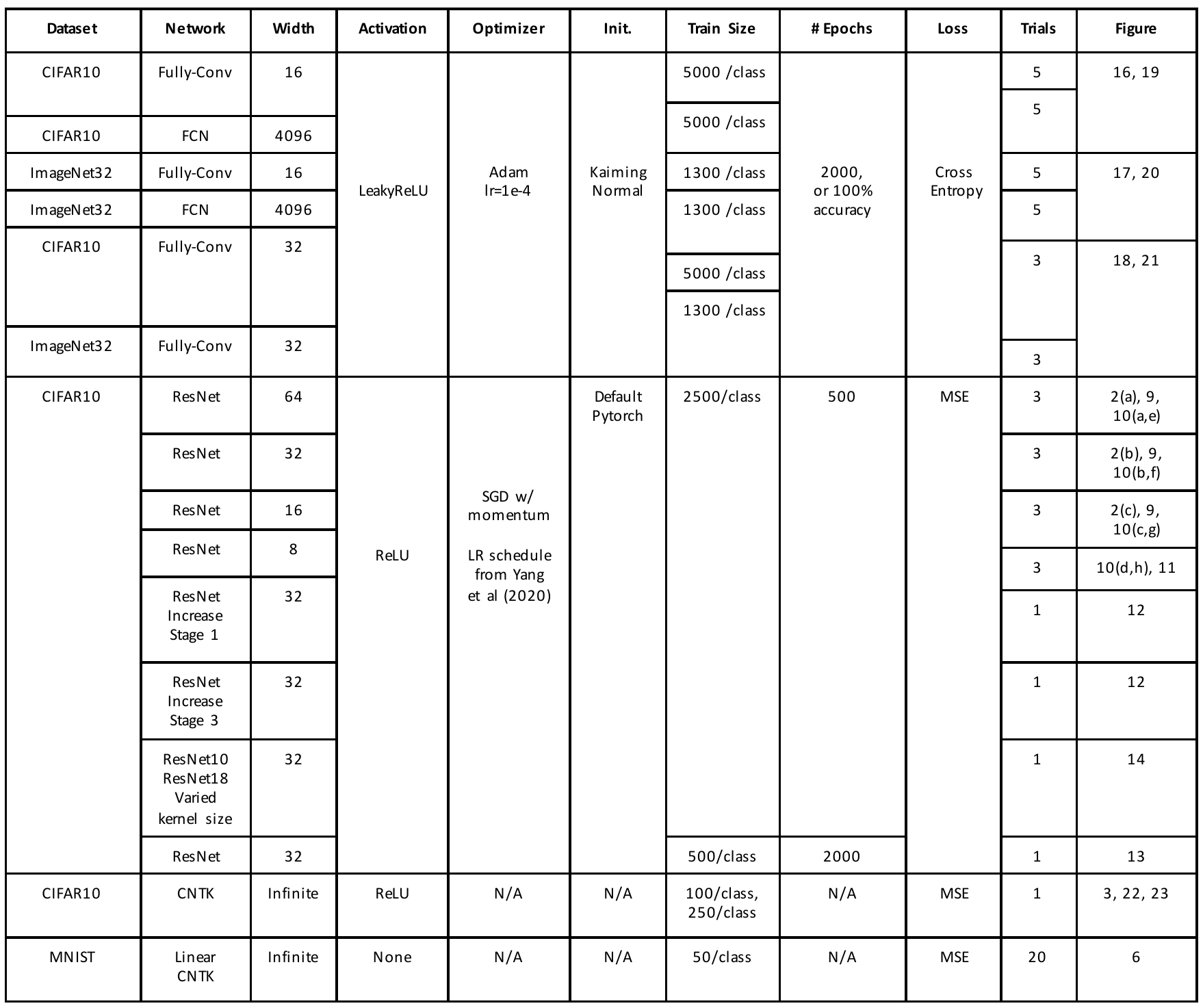}
    \caption{Experimental details for all experiments conducted.}
    \label{fig: all experimental details}
\end{figure*}
\clearpage
\begin{table}[h!]
\centering
\begin{tabular}{|c|c|}
    \hline
   \textbf{\# Classes} & \textbf{Classes used} \\
    \hline
    2 & dog, cat\\
    \hline
    5 & bird, cat, deer, dog, horse\\
    \hline
    10 & all classes\\
    \hline
\end{tabular}
\caption{\centering CIFAR10 classes considered in this work.}
\label{table: CIFAR10 classes}
\end{table}

\begin{table}[h!]
\centering
\begin{tabular}{|c|c|}
    \hline
    \textbf{\# Classes} & \textbf{Classes used} \\
    \hline
    2 & kit\_fox, English\_setter\\
    \hline
    5 & 2 classes + Siberian\_husky, Australian\_terrier, English\_springer\\
    \hline
    10 & 5 classes + Egyptian\_cat, Persian\_cat, malamute, Great\_Dane, Walker\_hound\\
    \hline
\end{tabular}
\caption{\centering ImageNet32 classes considered in this work.}
\label{table: imagenet32 classes}
\end{table}

\begin{table}[h!]
\centering
\begin{tabular}{|c|c|}
    \hline
    \textbf{Model Depth} & \textbf{Blocks per Stage} \\
    \hline
    10 & 1, 1, 1, 1\\
    \hline
    12 & 1, 1, 2, 1\\
    \hline
    14 & 1, 2, 2, 1\\
    \hline
    16 & 2, 2, 2, 1\\
    \hline
    18 & 2, 2, 2, 2\\
    \hline
    20 & 2, 2, 3, 2\\
    \hline
    22 & 2, 3, 3, 2\\
    \hline
    26 & 3, 3, 3, 3\\
    \hline
    30 & 3, 4, 4, 3\\
    \hline
    34 & 3, 4, 6, 3\\
    \hline
    38 & 3, 4, 8, 3\\
    \hline
    42 & 3, 4, 10, 3\\
    \hline
    46 & 3, 4, 12, 3\\
    \hline
    50 & 3, 4, 14, 3\\
    \hline
    60 & 3, 4, 19, 3\\
    \hline
    70 & 3, 4, 24, 3\\
    \hline
    80 & 3, 4, 29, 3\\
    \hline
    100 & 3, 4, 39, 3\\
    \hline
\end{tabular}
\caption{\centering Stage breakdown for all ResNet models used.}
\label{table: model depths}
\end{table}


\clearpage

\section{Additional ResNet Experiments}
\label{app: Additional Experiments} 

\subsection{ResNet Architecture}

\begin{figure}[!h]
    \centering
    \includegraphics[scale=0.4]{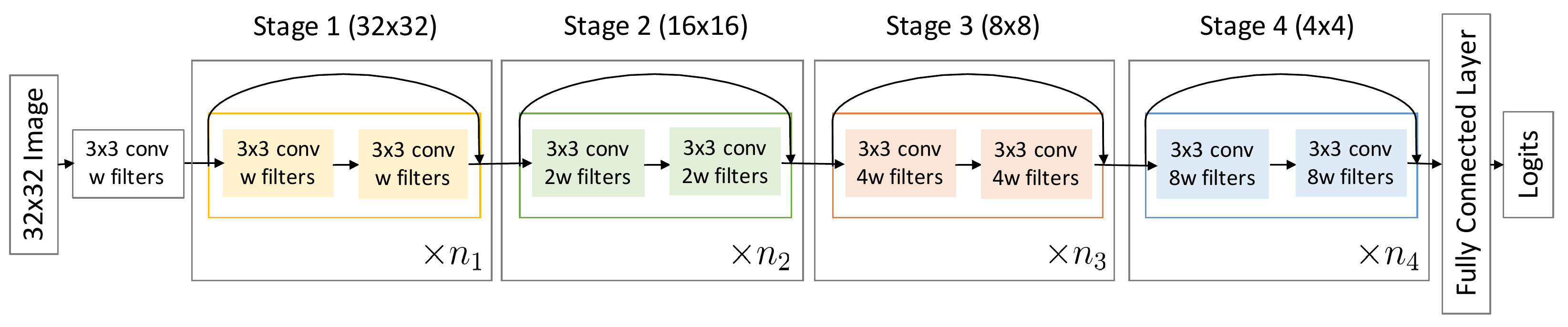}
    \vspace{-0.3cm}
    \caption{A diagram of the custom ResNet models used. The $i$th block is repeated $n_i$ times in stage~$i$. The first convolutional layer in stages 2, 3, and 4 downsamples the input using a stride of 2, while the remaining layers have a stride of 1.}
    \label{fig: custom ResNet}
\end{figure}

\subsection{Train and Test Losses, Accuracies}

\begin{figure*}[!h]
    \centering
    \begin{subfigure}[t]{.45 \textwidth}
        \centering
        \includegraphics[scale=0.3]{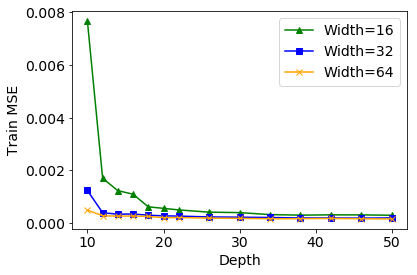}
        \caption{\centering Train Losses of ResNet models with widths 16, 32, and 64.}
    \end{subfigure}
    \begin{subfigure}[t]{.45 \textwidth}
        \centering
        \includegraphics[scale=0.3]{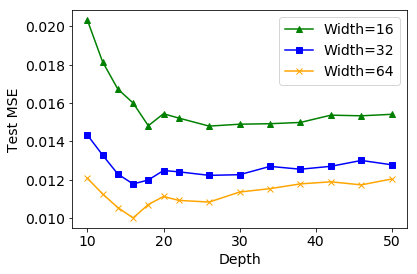}
        \caption{\centering Test Losses of ResNet models with widths 16, 32, and 64.}
    \end{subfigure}%
    \caption{\centering Train and Test losses of the ResNet models for all widths.}
    \label{fig: ResNet CIFAR10 All Widths}
\end{figure*}

\begin{figure*}[!h]
    \centering
    \includegraphics[scale=0.45]{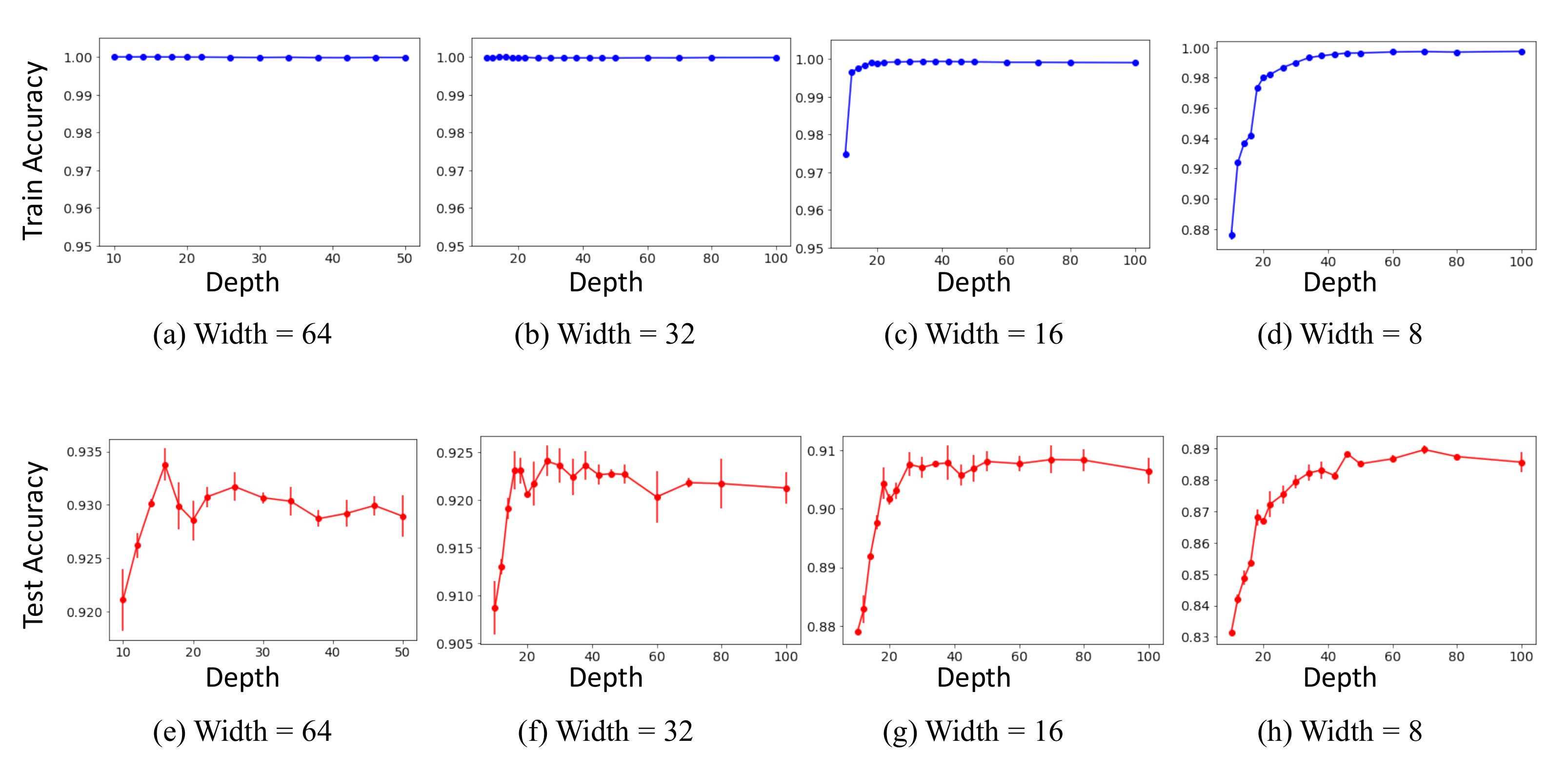}
    \caption{Train and Test accuracies for the ResNet models of width 8, 16, 32, and 64, for increasing depths.}
    \label{fig: resnet accuracies}
\end{figure*}

In Figure~\ref{fig: ResNet CIFAR10 All Widths}, we plot the train and test losses of all ResNet models used (for widths 16, 32, 64). Additionally, in Figure~\ref{fig: resnet accuracies} we plot the accuracies of all ResNet models.

\subsection{Effect of Depth in models with small widths}

\begin{figure*}[!h]
    \centering
    \begin{subfigure}[t]{.45 \textwidth}
        \centering
        \includegraphics[scale=0.4]{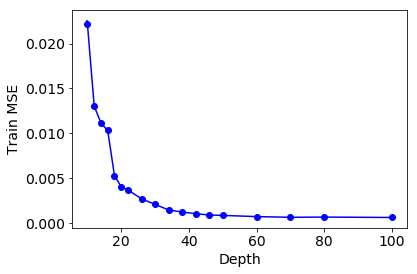}
        \caption{\centering Train losses for the width 8 ResNet model, for increasing depths.}
    \end{subfigure}
    \begin{subfigure}[t]{.45 \textwidth}
        \centering
        \includegraphics[scale=0.4]{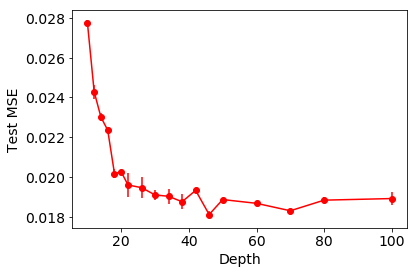}
        \caption{\centering Test losses for the width 8 ResNet model, for increasing depths.}
    \end{subfigure}
    \caption{Train and test losses for the width 8 ResNet model. We see that test loss decreases as model depth increases, but train loss has still not reached 0, even for large depths.}
    \label{fig: resnet width 8}
 \end{figure*}

In Figure~\ref{fig: resnet width 8}, we train ResNets of increasing depth using a smaller width (width 8). While the test loss doesn't clearly increase, it begins to plateau with sufficient depth. We argue this is because the width 8 model requires a much larger depth to reach zero training loss, while all the other models do at earlier depths.

\subsection{Effect of Downsampling}

\begin{figure*}[!h]
    \centering
    \includegraphics[scale=0.4]{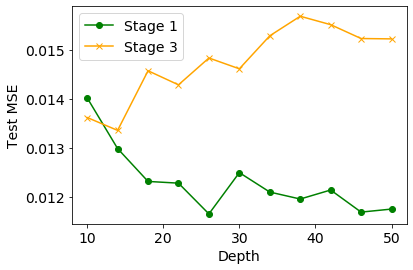}
    \caption{Test losses for the width 32 ResNet model where the 1st stage blocks are increased versus the model where the 3rd stage blocks are increased.}
    \label{fig: resnet stages}
\end{figure*}

In Figure~\ref{fig: resnet stages} we compare a ResNet model where we increase the number of blocks in the first stage versus a model where we increase the number of blocks in the third stage. We observe that the model where the third stage blocks are increased performs worse. This is likely because adding a block in a later stage, after downsampling, increases the effective depth of the model more than adding a block in an earlier stage.

\clearpage 
\subsection{Effect of Number of Samples}
\begin{figure*}[!h]
    \centering
    \includegraphics[scale=0.33]{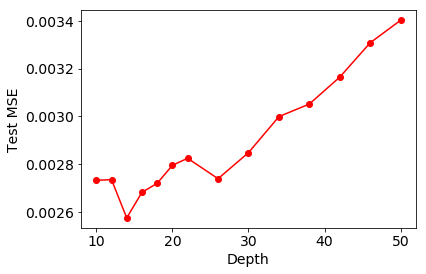}
    \caption{\centering Test losses for the width 32 ResNet samples, trained on 500 samples per class.}
    \label{fig: resnet less samples}
 \end{figure*}

In Figure~\ref{fig: resnet less samples} is a plot of test losses when training the width 32 ResNet model on 500 samples per class ($\frac{1}{10}$ of CIFAR10). Number of training epochs is increased accordingly. We observe that test loss increases as depth increases, showing that this phenomenon is robust to change in sample size.

\subsection{Effect of Kernel Size}

\begin{figure*}[!h]
    \centering
    \includegraphics[scale=0.45]{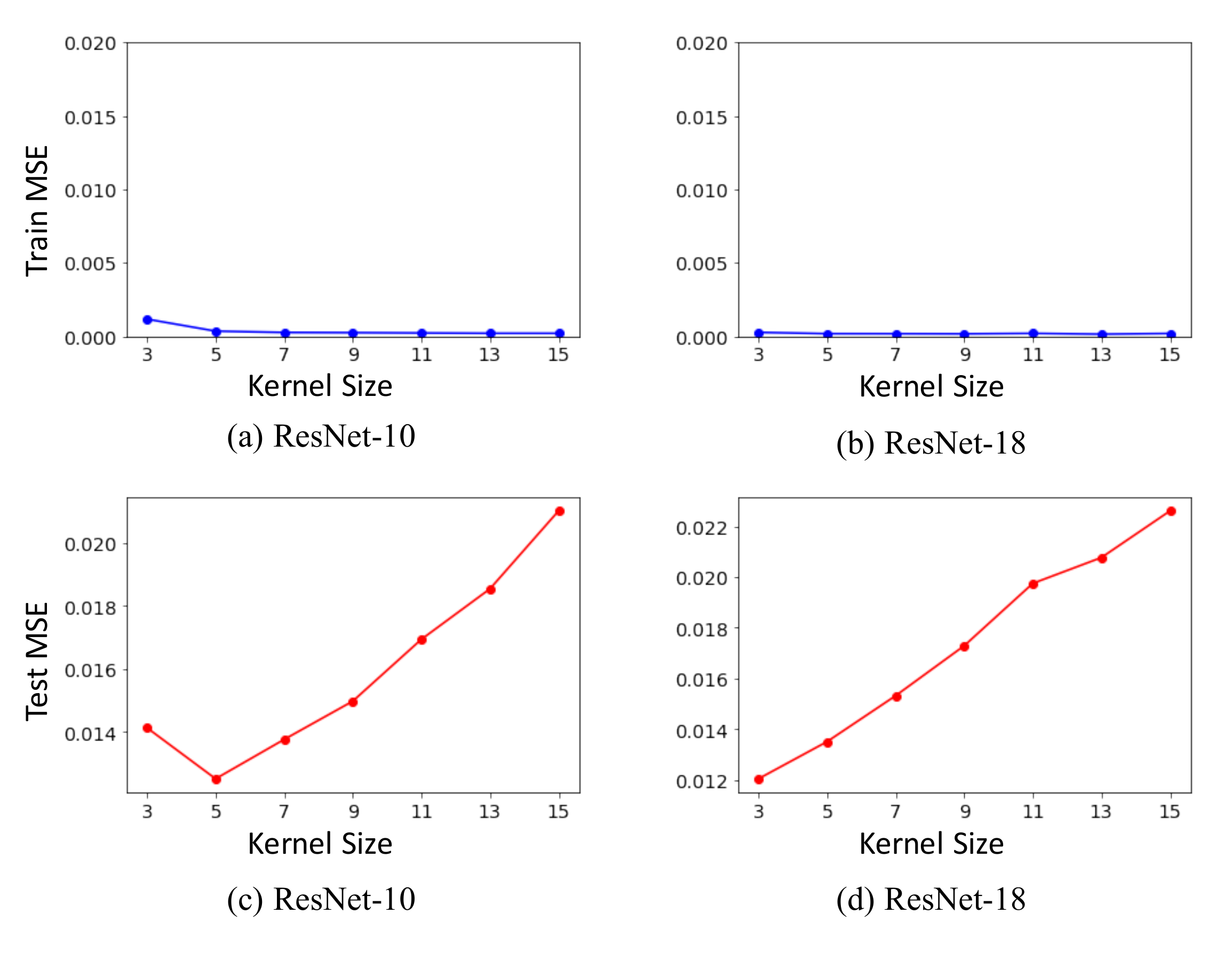}
    \caption{Train and test losses for width 32 models for increasing kernel size. We observe that test loss increases as kernel size increases.}
    \label{fig: resnet filter size}
 \end{figure*}

Another form of overparameterization is increasing the kernel size for convolutional filters. In Figure~\ref{fig: resnet filter size}, we train ResNet-10 and ResNet-18 of width 32 and varying kernel sizes, and observe that as kernel size increases, test loss increases. This is consistent with our proposed explanation based on expressivity, since increasing kernel size increases representational power independent of depth.

\clearpage

\section{Image Classification with Fully-Convolutional Networks}
\label{app: fully conv}

\begin{figure}[!h]
    \centering
    \includegraphics[scale=0.6]{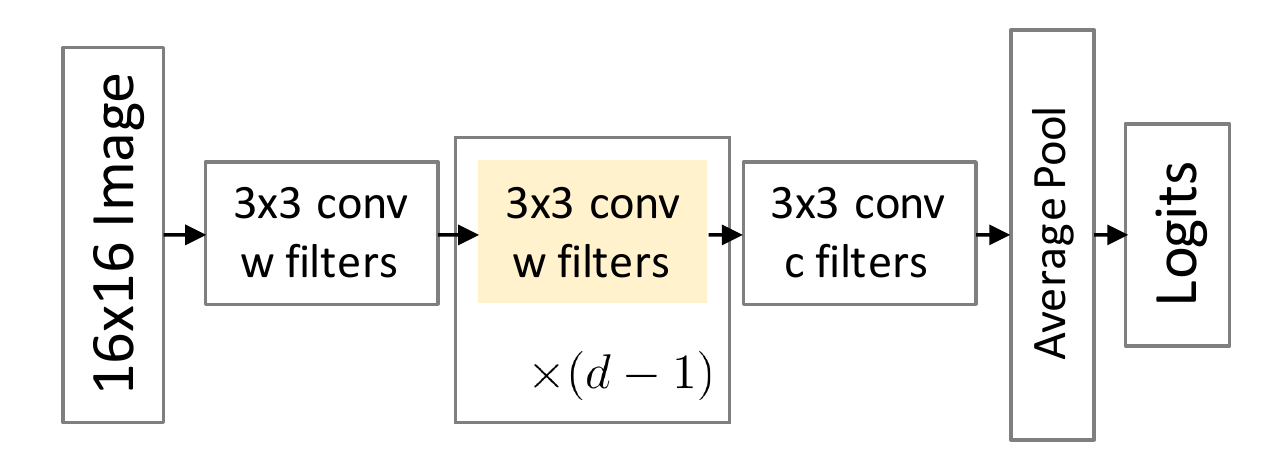}
    \caption{ A diagram of the Fully-Conv Net used with width $w$, depth $d$, and $c$ classes. Each convolutional layer (except for the last one) is followed by batch norm and LeakyReLU.}
    \label{fig: fully conv diagram}
\end{figure}

We also considered a simplified model of a convolutional network, which we call the \emph{Fully-Conv Net}. The architecture of a Fully-Conv Net of depth $d$ and width $w$ for a classification problem with $c$ classes is depicted in  Figure \ref{fig: fully conv diagram} and consists of the following layers:
\begin{itemize}
    \item A convolutional layer with stride 1, 3 input filters, and $w$ output filters, followed by batch norm~\cite{BatchNorm} and a LeakyReLU activation~\cite{LeakyReLU}.
    \item $d-1$ convolutional layers with stride 1, $w$ input filters, and $w$ output filters, each followed by batch norm and LeakyReLU activation.
    \item 1 convolutional layer with stride 1, $w$ input filters, and $c$ output filters. This is followed by an average pool of each of the output filters to produce a $c$-dimensional prediction.
\end{itemize}

Crucially, this network depends only on convolutional layers, a nonlinear activation, and batch norm; it does not depend on other components commonly found in deep learning architectures such as residual connections, dropout, downsampling, or fully connected layers. We note that this model is not designed to obtain optimal performance, but rather to isolate and understand the effect of increasing the number of convolutional layers.

\begin{figure*}[!h]
    \centering
    \includegraphics[scale=0.45]{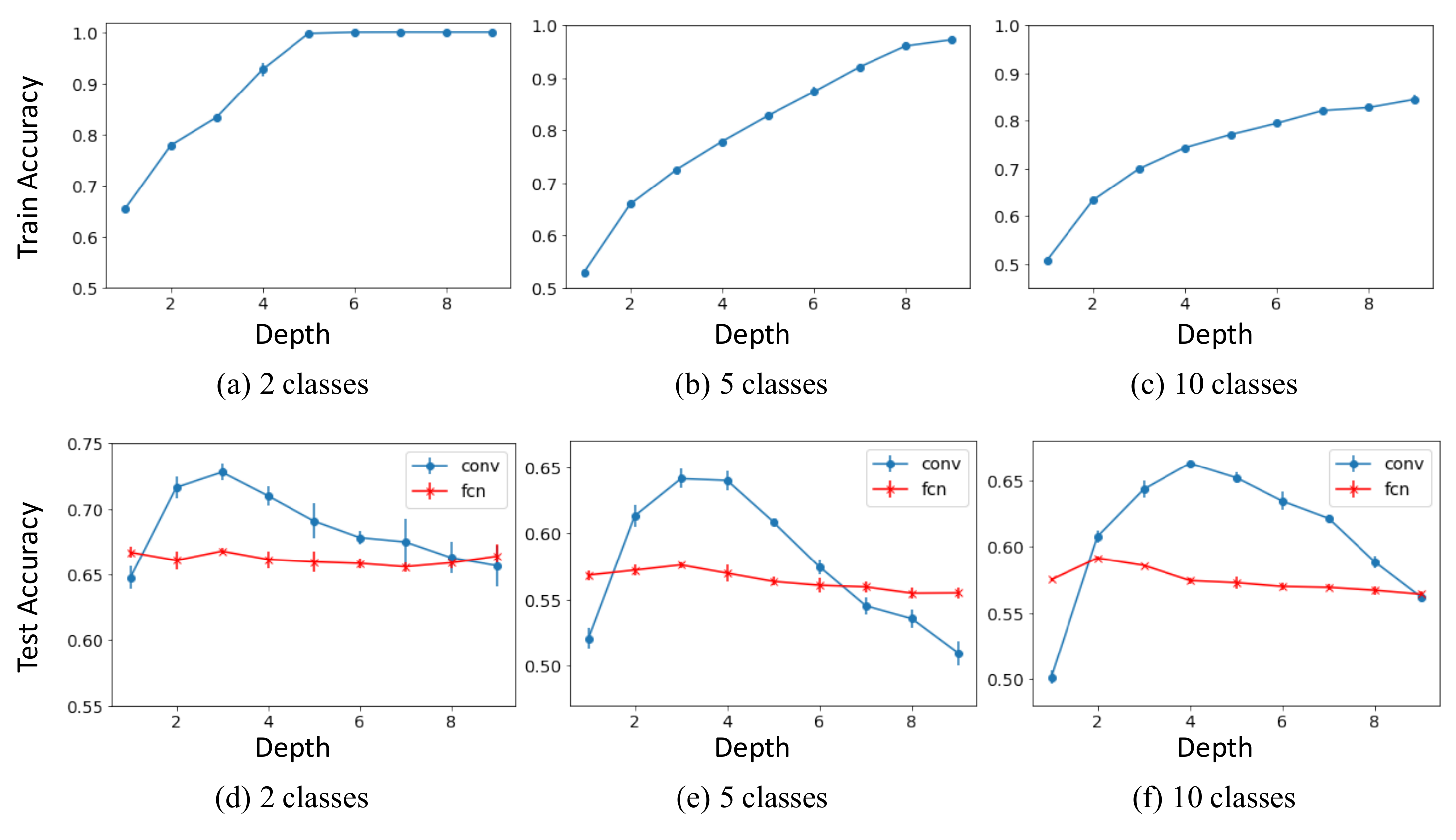}
    \caption{Train and test accuracy of the Fully-Conv Net as a function of depth, for CIFAR10 input images.  Increasing depth beyond a critical value leads to a decrease in test accuracy.  As depth increases, the performance of the Fully-Conv Net approaches that of a wide fully connected network (shown in red). All experiments are performed across 5 random seeds.}
    \label{fig: All Conv CIFAR10}
\end{figure*}

We trained the Fully-Conv Net on 2, 5, and 10 classes from CIFAR10. All experiments were performed using 5 random seeds to reduce the impact of random initialization. Models were trained using Adam \cite{Adam} with learning rate $10^{-4}$ for 2000 epochs, and we selected the model with the best training accuracy over the course of training. We used the Cross Entropy loss, and down-sampled images to $16 \times 16$ resolution to reduce the computational burden. See Appendix~\ref{app: experiments} for a list of all classes used. The resulting train and test accuracies are shown in Figure~\ref{fig: All Conv CIFAR10}. As expected, as depth increases, training accuracy becomes 100\%. However, beyond a critical depth threshold, the test accuracy begins to degrade sharply. 

\begin{figure*}[!h]
    \centering
    \includegraphics[scale=0.45]{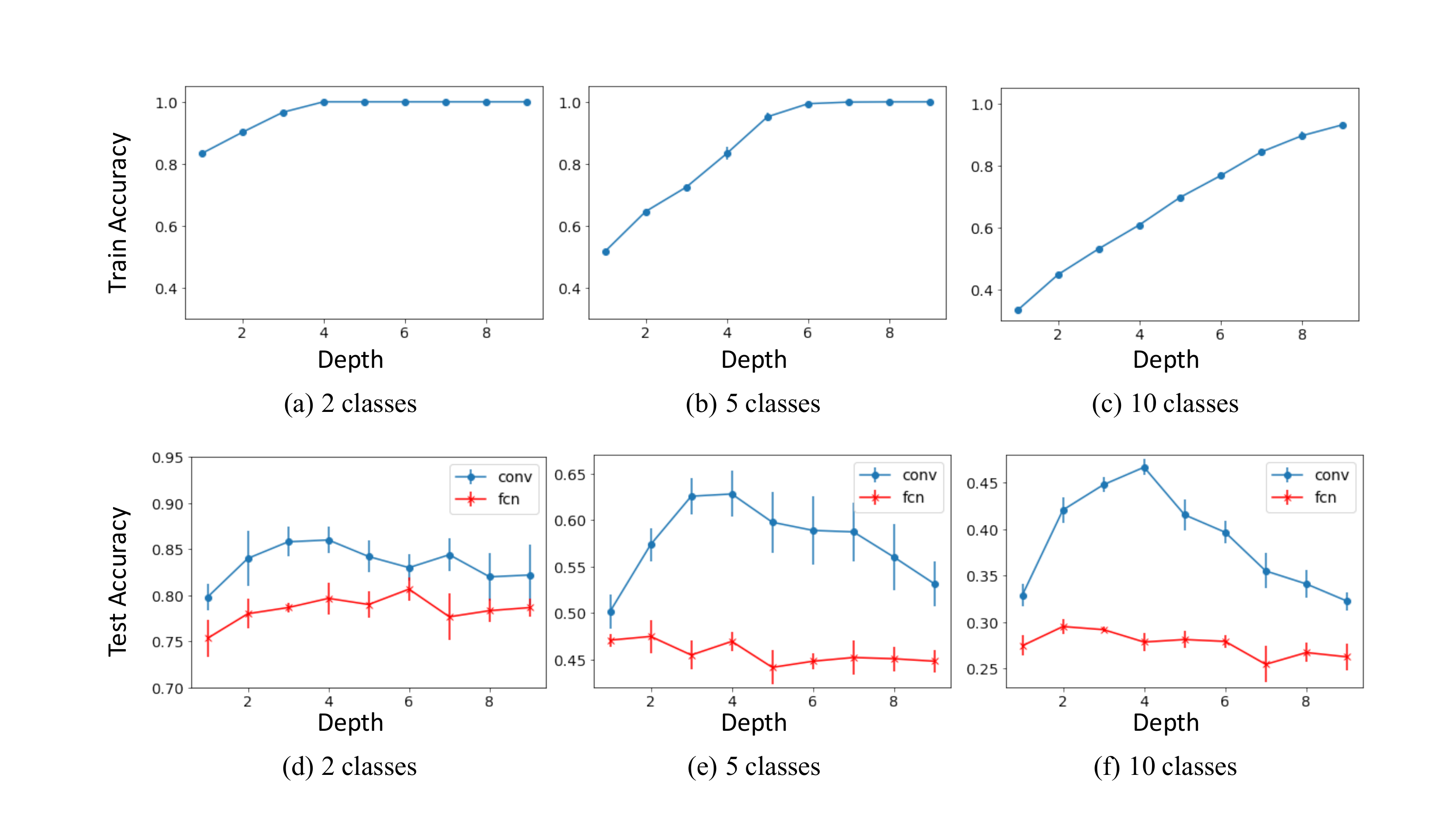}
    \caption{Train and test accuracy of the Fully-Conv Net as a function of depth, for ImageNet32 input images. Increasing depth beyond a critical threshold again leads to a decrease in test accuracy.  Increasing depth beyond a critical value leads to a decrease in test accuracy.  As depth increases, the performance of the Fully-Conv Net approaches that of a wide fully connected network (shown in red). All experiments are performed across 5 random seeds. }
    \label{fig: All Conv ImageNet32}
\end{figure*}

In addition to CIFAR10, we also applied the Fully-Conv Net to subsets of ImageNet32~\cite{ImageNet32}, which is  ImageNet downsampled to size $32 \times 32$. We again trained on 2, 5, and 10 classes, using the same training procedure as for CIFAR10. Training and test accuracies for ImageNet32 are shown in Figure~\ref{fig: All Conv ImageNet32}. Again, we observe that as depth increases past a critical value, test performance degrades.

\clearpage

\begin{figure*}[!h]
    \centering
    \includegraphics[scale=0.4]{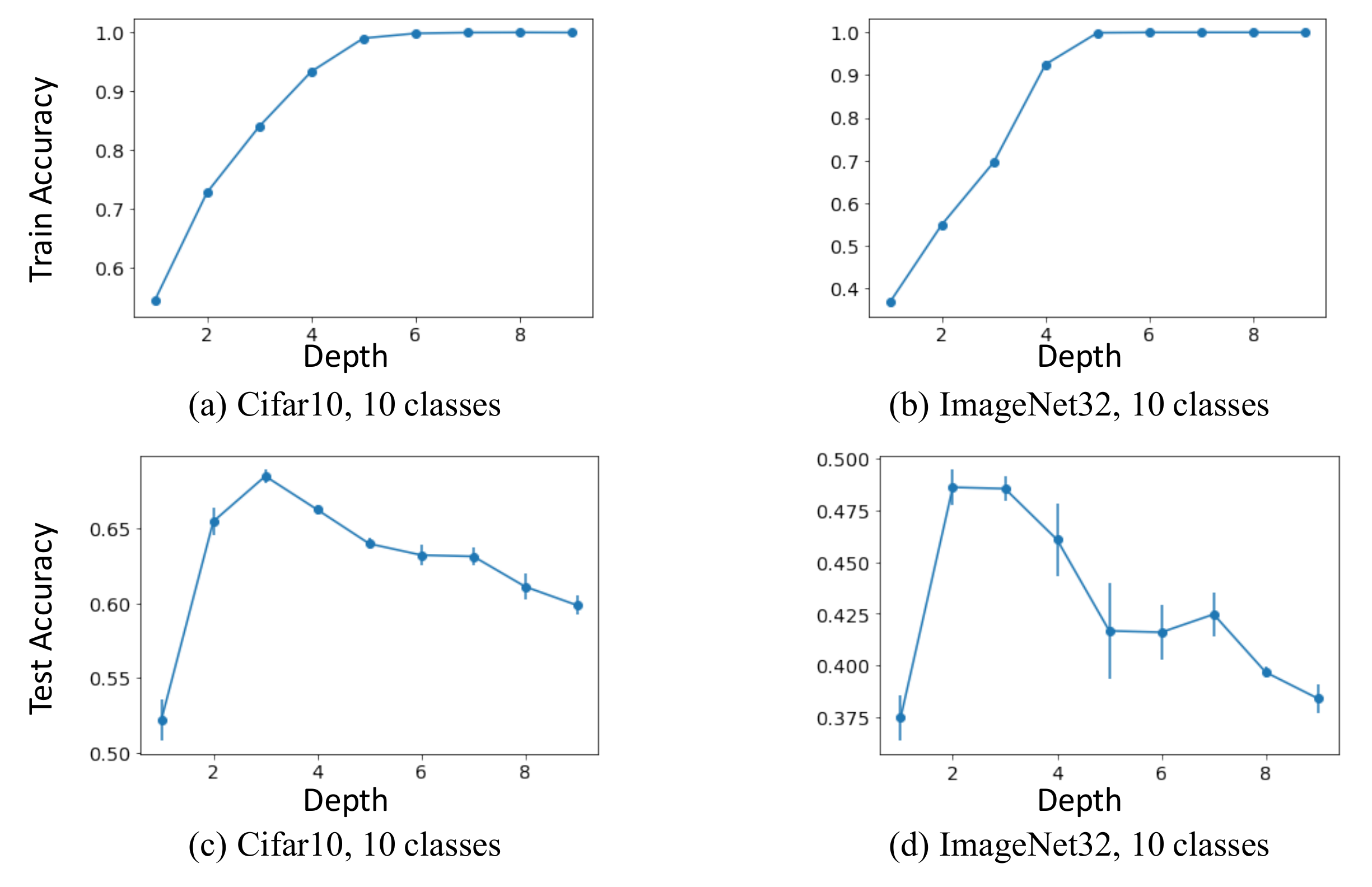}
    \caption{Increasing depth well past the interpolation threshold leads to a decrease in test accuracy.  All experiments are across 3 random seeds. (a) We trained Fully-Conv nets of increasing depth but fixed width 32 on 10 classes of CIFAR10. (b) We trained Fully-Conv nets of increasing depth but fixed width 32 on 10 classes of ImageNet32.}
    \label{fig: wider models}
\end{figure*}

When training to classify between 2 and 5 classes, the test accuracy continues to decrease even when increasing depth past the interpolation threshold, i.e. even after achieving $100\%$ training accuracy.  This in contrast to double descent where increasing model complexity beyond the interpolation threshold leads to an increase in test accuracy.  Interestingly, as depth increases, the test accuracy approaches and can even be worse than that of a fully connected network.  While the Fully-Conv Nets were before or at the interpolation threshold for the 10 class setting in Figures \ref{fig: All Conv CIFAR10} and \ref{fig: All Conv ImageNet32}, in Figure~\ref{fig: wider models} in we demonstrate that a similar decrease in test accuracy occurs also after the interpolation threshold for wider models which can interpolate the data.

\begin{figure*}[!h]
    \centering
    \includegraphics[scale=0.45]{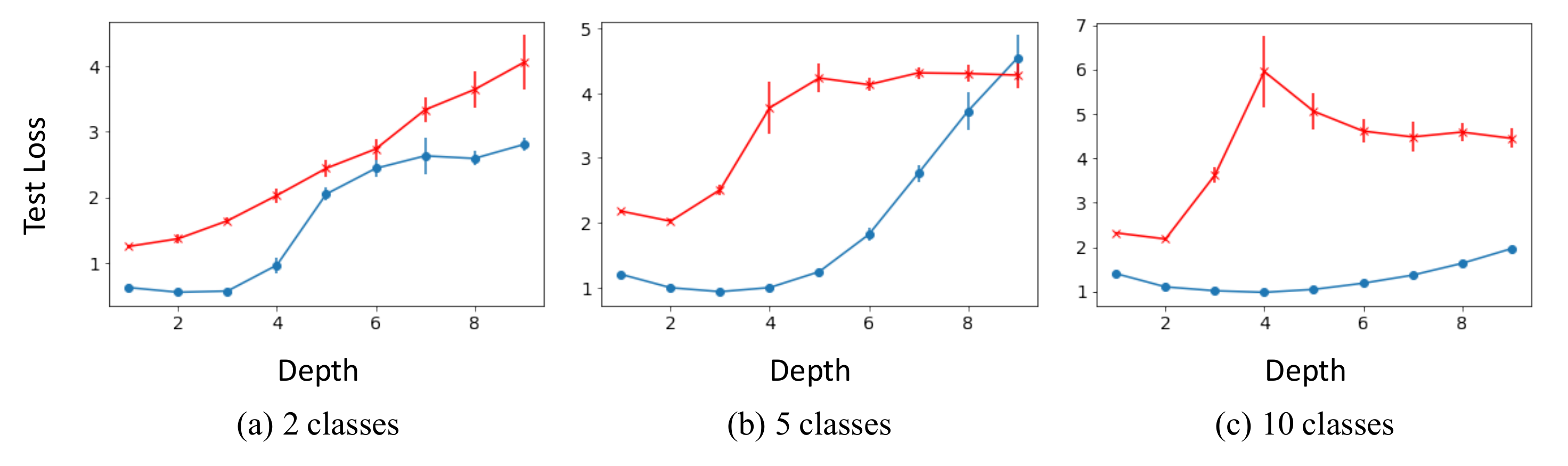}
    \caption{Test losses for models in Figure 2 (on CIFAR10). The error for the convolutional networks is in blue and that of full connected networks is in red.}
    \label{fig: fully conv cifar loss}
\end{figure*}

\begin{figure*}[!h]
    \centering
    \includegraphics[scale=0.45]{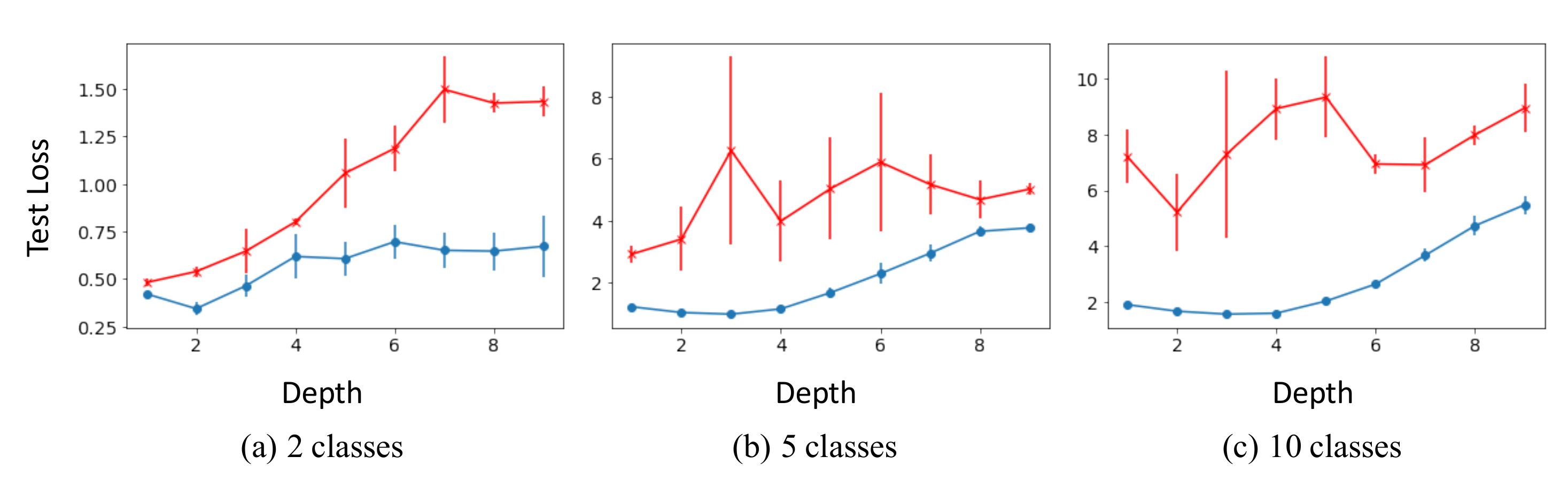}
    \caption{Test losses for models in Figure 3 (on ImageNet32). The error for the convolutional networks is in blue and that of full connected networks is in red.}
    \label{fig:fully conv imagenet losses}
\end{figure*}

\begin{figure*}[!h]
    \centering
    \includegraphics[scale=0.45]{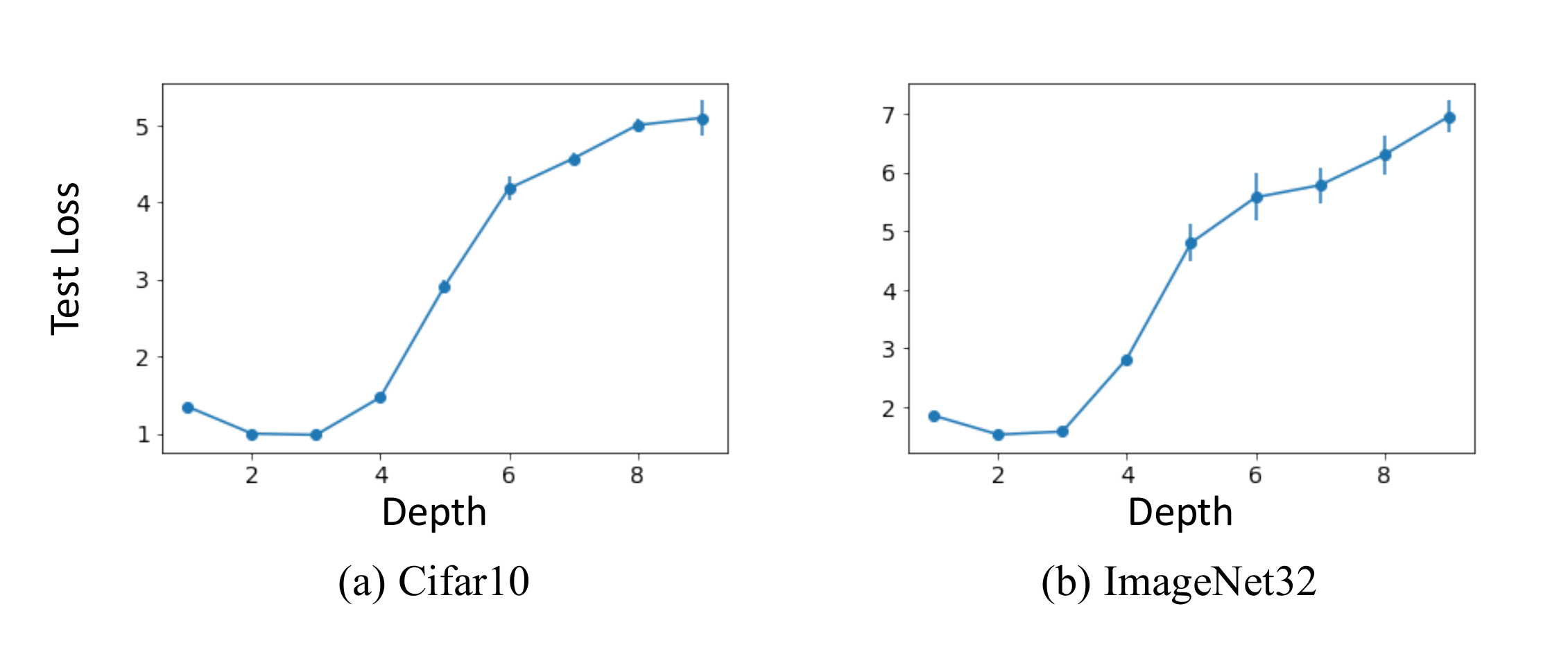}
    \caption{Test losses for the models in Figure~\ref{fig: wider models}.}
    \label{fig: wide losses}
\end{figure*}

We also plot the test cross entropy loss of all fully convolutional models trained on CIFAR10 and ImageNet32. Figure~\ref{fig: fully conv cifar loss} contains the test losses for the models trained on CIFAR10, Figure~\ref{fig:fully conv imagenet losses} contains the test losses for the models trained on ImageNet32, and Figure~\ref{fig: wide losses} contains the test losses for the wider models considered in Figure~\ref{fig: wider models}. In all cases, we observe that test loss begins to increase beyond a critical depth.

\clearpage

\clearpage 
\section{Additional CNTK Experiments}
\label{sec: CNTK extra}

\begin{figure*}[!h]
    \centering
    \includegraphics[scale=0.45]{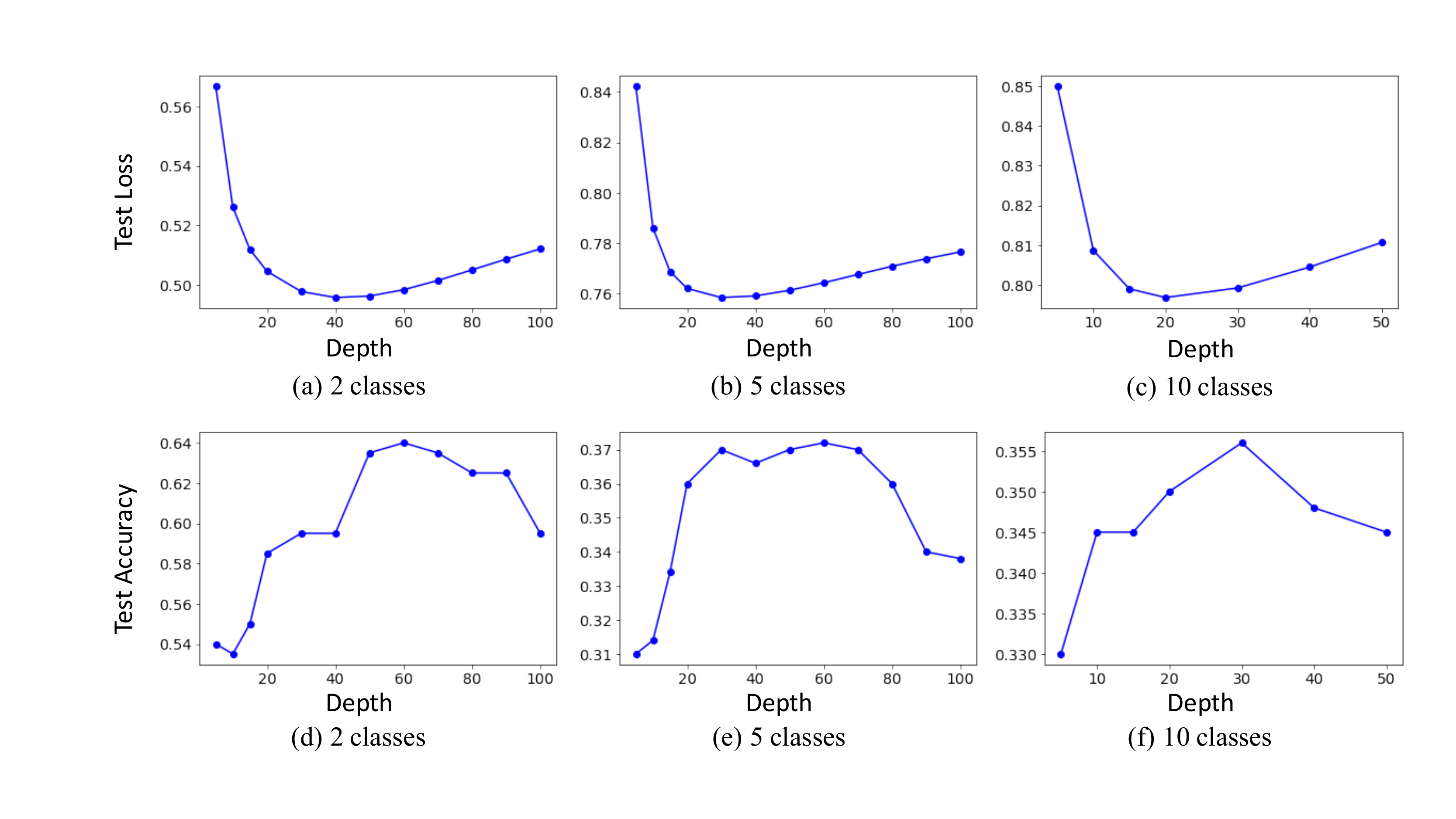}
    \caption{Test loss and test accuracy for the CNTK trained on subsets of CIFAR10 with 2, 5 and 10 classes. We observe that generalization improves up to a critical depth, after which it worsens.}
    \label{fig: CNTK classes}
 \end{figure*}

We also train the CNTK on subsets of CIFAR10 of varying number of classes. We use 100 train and 100 test examples per class, and train on both 2 classes (birds and deer), 5 classes (cats, dogs, horses, bids, and deer), and 10 classes (all of CIFAR10). The test losses and accuracies are shown in Figure~\ref{fig: CNTK classes}. Again, we see that generalization is unimodal, with test loss decreasing until a critical depth and increasing afterwards, which is in agreement with our main CNTK experiment in Figure~\ref{fig: CNTK}.

We note that training the CNTK for large depths is computationally prohibitive. The runtime scales quadratically in the number of training samples; furthermore, training the depth 500 CNTK on 1 GPU for 500 train and 500 test samples took approximately 2 days.

\begin{figure*}[!h]
    \centering
    \includegraphics[scale=0.45]{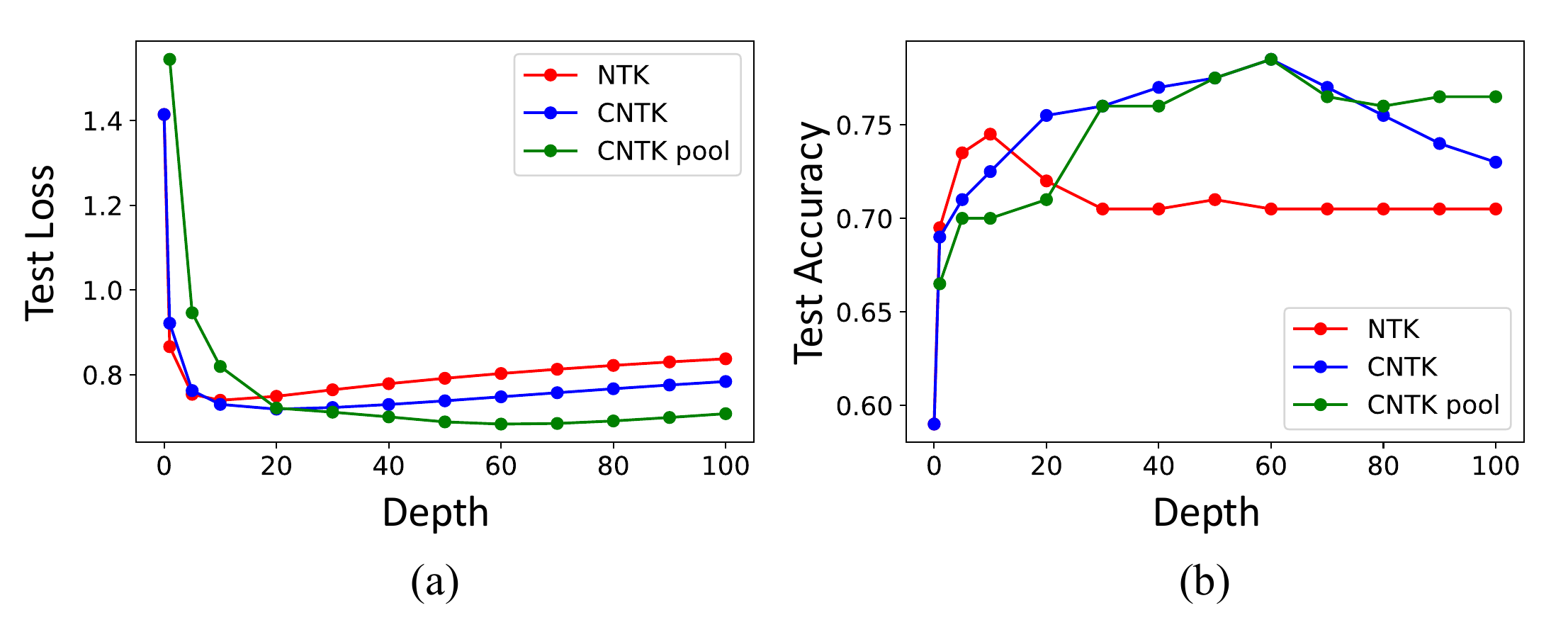}
    \caption{Test loss and test accuracy for the NTK, CNTK, and CNTK with pooling trained on 2 classes of CIFAR10.}
    \label{fig: CNTK with pooling}
 \end{figure*}

We also analyze the test loss of the CNTK where the last layer is a global average pooling layer. The test loss and test accuracy for the models trained on 2 classes of CIFAR10 (with 100 train samples per class) are shown in Figure~\ref{fig: CNTK with pooling}. Even with the CNTK with pooling we observe the similar trend of decreasing loss, and then increasing loss past a critical depth. We use Neural Tangents~\cite{neuraltangents2020} for this experiment; computing the CNTK with pooling for larger depths on more samples was computationally prohibitive.

\clearpage 

\section{Proofs} \label{app: proofs}

\subsection{Proof of Theorem 1}

\begin{proof}
For a depth $d \le D$ in the network, define $\mathcal{K}^{(d)}$, $\Gamma^{(d)}$ to be the Neural Network Gaussian Process (NNGP) and NTK respectively at layer $d$. Concretely, for inputs $x, y \in \mathbb{R}^p$, define
\begin{align}
    \mathcal{K}^{(d)}(x, y) &= \mathbb{E}\left[x_i^{(d)}{y_i^{(d)}}^T\right] \in \mathbb{R}^{p \times p}\\
    \Gamma^{(d)}(x, y) &= \mathbb{E}\left[\frac{\partial x_i^{(d)}}{\partial \bm{\theta}^{\le d}}\left( \frac{\partial {y}_i^{(d)}}{\partial \bm{\theta}^{\le d}}\right)^T \right] \in \mathbb{R}^{p \times p},
\end{align}
where $\bm{\theta}^{\le d}$ is the vector of parameters up to and including layer $d$. It is straightforward to compute the NNGP -- since the $\theta$ are all i.i.d $\mathcal{N}(0, 1)$, and the values of each channel at a given depth $x_i^{(d)}$ are also i.i.d Gaussians, plugging in the recursion from \eqref{eq: CNTK def} we obtain
\begin{align}
    \mathbb{E}\left[x_i^{(d)}{{y}_i^{(d)}}^T\right] &= \sum_{k=1}^r \mathbb{E}\left[(\theta_{ij; k}^{d})^2B_kx_j^{(d-1)}{{y}_j^{(d-1)}}^TB_k^T\right]\\
    &= \sum_{k=1}^r B_k \mathbb{E}\left[x_j^{(\ell-1)}{{y_j}^{(\ell-1)}}^T\right] B_k^T.
\end{align}
This gives the recursion
\begin{equation} \label{eq: NNGP recursion}
    \mathcal{K}^{(d)}(x,y) = \sum_{k=1}^r B_k\mathcal{K}^{(d-1)}(x,y)B_k^T = \mathcal{A}(\mathcal{K}^{(d-1)}(x,y))
\end{equation}
by definition of $\mathcal{A}$. The NTK recursion can be calculated in a similar manner to the CNTK computation in~\cite{XiaoGeneralization}; since the activations are linear and have derivative 1, using \cite[Equation 152]{XiaoGeneralization} we obtain
\begin{equation}
\Gamma^{(d)}(x, y) = \mathcal{K}^{(d)}(x,y) + \mathcal{A}(\Gamma^{(d - 1)}(x, y))
\end{equation}
This formula tells us that the NTK is a constant (depending on $d$) times the NNGP, so we only need to focus on how the NNGP evolves. The initial condition for the NNGP is $\mathcal{K}^{(0)}(x, y) = x{y}^T$, so substituting into \eqref{eq: NNGP recursion} and unrolling the recursion gives
\begin{equation} \label{eq: NNGP unrolled}
    \mathcal{K}^{(d)}(x, y) = \sum_{k_1, \dots, k_\ell \in [r]^{d}}B_{k_d}\cdots B_{k_1} x {y}^TB_{k_1}^T \cdots B_{k_d}^T.
\end{equation}

Let us first consider a pooling network $f^D_{pool}(x; \bm{\theta})$ with CNTK between two inputs $x, y$ being $K^D_{pool}(x, y) \in \mathbb{R}$. We have that 
\begin{equation}
    K^{D}_{pool}(x, y) = \frac{1}{p^2}\sum_{i,j}\Gamma^{(D)}(x, y)_{ij}.
\end{equation}
Plugging in \eqref{eq: NNGP unrolled}, we get
\begin{align}
K^{D}_{pool}(x, y) &= c\sum_{k_1, \dots, k_D \in [r]^{D}} \sum_{i,j}(B_{k_D}\cdots B_{k_1} x {y}^TB_{k_1}^T \cdots B_{k_D}^T)_{ij}\\
&= c\sum_{k_1, \dots, k_D \in [r]^{D}} \sum_{i,j}((B_{k_D}\cdots B_{k_1} x) (B_{k_D}\cdots B_{k_1} y)^T)_{ij}\\
&= c\sum_{k_1, \dots, k_D \in [r]^{D}} (B_{k_D}\cdots B_{k_1} x)^T \mathbf{1} \mathbf{1}^T(B_{k_D}\cdots B_{k_1} y)\\
&= x^T\left(c\sum_{k_1, \dots, k_D \in [r]^{D}} B_{k_1}^T \cdots B_{k_D}^T\mathbf{1} \mathbf{1}^TB_{k_D}\cdots B_{k_1}\right)y
\end{align}
where $c$ is a constant independent of $x, y$. After rescaling, we see that we can write the kernel as
$$K^D_{pool}(x, y) = x^T \Theta_D y,$$ where $\Theta_0 = \mathbf{1} \mathbf{1}^T = J_p$ and $\Theta_n$ satisfies the recursion
$$\Theta_D = \sum_{k \in [r]}B_k^T\Theta_{D-1}B_k = \mathcal{A}(\Theta_{D-1}),$$ as desired.
Next, consider a flattening network. The CNTK between two inputs $K^D_{flat}(x, y)$ is given by
\begin{equation}
    K^{D}_{flat}(x, y) = \frac{1}{p}\sum_{i}\Gamma^{(D)}(x, y)_{ii}.
\end{equation}
Thus
\begin{align}
K^D_{flat}(x, y) &= c\sum_{k_1, \dots, k_L \in [r]^{D}} \sum_{i}(B_{k_D}\cdots B_{k_1} x {x}^TB_{k_1}^T \cdots B_{k_D}^T)_{ii}\\
&= c\sum_{k_1, \dots, k_D \in [r]^{D}} \sum_{i}((B_{k_D}\cdots B_{k_1} x) (B_{k_D}\cdots B_{k_1} y)^T)_{ii}\\
&= c\sum_{k_1, \dots, k_D \in [r]^{D}} (B_{k_D}\cdots B_{k_1} x)^T (B_{k_D}\cdots B_{k_1} y)\\
&= x^T\left(c\sum_{k_1, \dots, k_D \in [r]^{D}} B_{k_1}^T \cdots B_{k_D}^TB_{k_D}\cdots B_{k_1}\right)y
\end{align}
Similarly to before, we can write the kernel as $K^D_{flat}(x, y) = x^T \Theta_D y$, where $\Theta_D$ satisfies the same recursion, $\Theta_D = \mathcal{A}(\Theta_{D-1})$, and the initial condition is now $\Theta_0 = I_p$.
\end{proof}

\subsection{Proof of Theorem 2}
\label{appendix: Proof of Theorem 2}

\begin{proof}
By the bias-variance decomposition, we can write 
\begin{equation}
    \mathcal{R}(\Theta; \beta, X) = \mathcal{B}(\Theta; \beta, X) + \mathcal{V}(\Theta; \beta, X),
\end{equation}
where the squared bias and variance conditioned on $X$ are
\begin{align}
    \mathcal{B}(\Theta; \beta, X)&= \mathbb{E}_x\left[(x^T\beta - \mathbb{E}_Y\hat{f}(x))^2\right]\\
    \mathcal{V}(\Theta; \beta, X)&= \mathbb{E}_x\left[(\hat{f}(x) - \mathbb{E}_Y\hat{f}(x))^2\right]
\end{align}
First we compute the variance. 
\begin{equation}
    \mathbb{E}_Y\hat{f}(x) = x^T\Theta X^T(X\Theta X^T)^{-1}\mathbb{E}_YY = x^T\Theta X^T(X\Theta X^T)^{-1}X\beta,
\end{equation}so
\begin{align}
    \mathcal{V}(\Theta; \beta, X) &= \mathbb{E}_x\left[(x^T\Theta X^T(X\Theta X^T)^{-1}\eta)^2 \right]\\
    &= \mathbb{E}_x\left[\eta^T(X\Theta X^T)^{-1}X\Theta xx^T\Theta X^T(X\Theta X^T)^{-1}\eta \right]\\
    &= \sigma^2 \text{Tr}\left[(X\Theta X^T)^{-1}X\Theta\Sigma\Theta X^T(XBX^T)^{-1}\right]\\
    &= \sigma^2\text{Tr}\left[(X\Theta X^T)^{-2}X\Theta\Sigma\Theta X^T\right].
\end{align}
We do a change of basis. Write $X = Z\Sigma^{1/2}$ ($x = \Sigma^{1/2}z$, where $z \sim \mathcal{N}(0, I)$). Then, letting $\Tilde{\Sigma} = \Sigma^{1/2}\Theta\Sigma^{1/2}$, the variance is
\begin{equation}
    \mathcal{V}(\Theta; \beta, X) = \sigma^2\text{Tr}\left[(Z\Tilde{\Sigma}Z^T)^{-2}Z\Tilde{\Sigma}^2Z^T\right]
\end{equation}
Therefore the expected variance over the randomness of the data is 
\begin{equation}
   \mathcal{V}(\Theta) = \sigma^2\mathbb{E}_{z_i \sim \mathcal{N}(0, I)}\text{Tr}\left[(Z\Tilde{\Sigma}Z^T)^{-2}Z\Tilde{\Sigma}^2Z^T\right], 
\end{equation}
where the entries of $Z$ are all i.i.d standard normal.

Next we consider the bias. We have
\begin{align}
    \mathcal{B}(\Theta; \beta, X) &:= \mathbb{E}_x\left[(x^T\beta - x^T\Theta X^T(X\Theta X^T)^{-1}X\beta )^2\right]\\
    &= \mathbb{E}_x\left[(x^T(I - \Theta X^T(X\Theta X^T)^{-1}X)\beta)^2\right]\\
    &:= \mathbb{E}_x\left[(x^TP_{\Theta; X}^{\perp}\beta)^2\right]\\
    &= \|P_{\Theta; X}^{\perp}\beta\|_{\Sigma}^2
\end{align}
Therefore 
\begin{equation}
    \mathcal{B}(\Theta; \beta) = \mathbb{E}_X \|P_{\Theta; X}^{\perp}\beta\|_{\Sigma}^2,
\end{equation}
as desired.
\end{proof}

\subsection{Proof of Proposition~\ref{prop: variance}}
\label{appendix: Proof of Prop 1}

\begin{proof}
For $Z$ such that $ZZ^T$ and $Z\Tilde{\Sigma}Z^T$ are invertible (which occurs almost surely) we have that
\begin{align}
   \text{Tr}\left[(Z\Tilde{\Sigma}Z^T)^{-2}Z\Tilde{\Sigma}^2Z^T\right]\text{Tr}[(ZZ^T)^{-1}] &= \text{Tr}\left[(Z\Tilde{\Sigma}Z^T)^{-1}Z\Tilde{\Sigma}^2Z^T(Z\Tilde{\Sigma}Z^T)^{-1}\right]\text{Tr}[(ZZ^T)^{-1}ZZ^T(ZZ^T)^{-1}]\\
   &\ge \text{Tr}\left[(Z\Tilde{\Sigma}Z^T)^{-1}Z\Tilde{\Sigma}Z^T(ZZ^T)^{-1}\right]\\
   &= \text{Tr}\left[(ZZ^T)^{-1}\right]^2,
\end{align}
where the inequality is due to Cauchy-Schwarz on the matrix inner product $\langle A, B \rangle = \text{Tr}(A^TB)$.
Therefore we can lower bound the variance by
\begin{equation}
V = \sigma^2\mathbb{E}[\text{Tr}\left[(Z\Tilde{\Sigma}Z^T)^{-2}Z\Tilde{\Sigma}^2Z^T\right]] \ge \sigma^2\mathbb{E}\text{Tr}\left[(ZZ^T)^{-1}\right] = \sigma^2\frac{n}{p - n - 1},
\end{equation}
since $(ZZ^T)^{-1}$ follows an inverse Wishart distribution with $p$ degrees of freedom. By the Cauchy-Schwarz equality condition, equality occurs if and only if $(Z\Tilde{\Sigma}Z^T)^{-1}Z\Tilde{\Sigma} = c(ZZ^T)^{-1}Z$ with probability 1. Multiplying both sides by $Z^T$ gives $c = 1$. Thus we must have, with probability 1:
\begin{align}
    (Z\Tilde{\Sigma}Z^T)^{-1}Z\Tilde{\Sigma} &= (ZZ^T)^{-1}Z\\
    \Longrightarrow Z\Tilde{\Sigma} &= Z\Tilde{\Sigma}Z^T(ZZ^T)^{-1}Z\\
    \Longrightarrow Z\Tilde{\Sigma}P_{Z}^\perp &= 0,
\end{align}
where $P_{Z}^\perp$ is the projection on $null(Z)$, which is nonzero since $n < p$. Therefore $\Tilde{\Sigma}null(Z) \subset null(Z)$ This can only occur if $null(Z) = span(v_1, \dots, v_{p-n})$ for some eigenvectors $v_1, \dots, v_{p-n}$ of $\Tilde{\Sigma}$. However, the only way this can be true with probability 1 over the choices of $Z$ is if all vectors in $\mathbb{R}^p$ are eigenvectors of $\Tilde{\Sigma}$, i.e. $\Tilde{\Sigma} = cI$ for a constant $c$. Since $\Tilde{\Sigma} = \Sigma^{\frac12}\Theta\Sigma^{\frac12}$, we thus have $\Theta = c\Sigma^{-1}$, which proves our desired claim.
\end{proof}

\subsection{Proof of Proposition~\ref{prop: bias}}
\label{appendix: Proof of Prop 2}

\begin{proof}
Since scaling $\Theta$ does not change the predictions, it suffices to show that predicting using the transformation $\Theta + \beta\beta^T$ has a smaller bias than that of using the transformation $\Theta$.
By the Woodbury identity, we have that
\begin{align}
   \left(X(\Theta + \beta\beta^T)X^T\right)^{-1} &= (X\Theta X^T + X\beta\beta^TX^T)^{-1}\\
   &= (K + uu^T)^{-1}\\
   &= K^{-1} - \frac{K^{-1}uu^TK^{-1}}{1 + u^TK^{-1}u}, 
\end{align}
where we've let $u := X\beta$ and $K := X\Theta X^T$. Therefore (setting $z = u^TK^{-1}u$ for convenience)
\begin{align}
    P_{\Theta + \beta\beta^T; X}\beta &= \left(I - (\Theta + \beta\beta^T)X^T\left(X(\Theta + \beta\beta^T)X^T\right)^{-1}X\right)\beta\\
    &= \beta - (\Theta + \beta\beta^T)X^T\left(K^{-1} - \frac{K^{-1}uu^TK^{-1}}{1 + z}\right)X\beta\\
    &= \beta - \Theta X^TK^{-1}X\beta - \beta\beta^TX^TK^{-1}X\beta + \frac{\Theta X^TK^{-1}uu^TK^{-1}u}{1 + z} + \frac{\beta uK^{-1}uu^TK^{-1}u}{1+z}\\
    &= \beta - \Theta X^TK^{-1}X\beta - z\beta + \Theta X^TK^{-1}X\beta\frac{z}{1 + z} + \frac{z^2}{1 + z}\beta\\
    &= \frac{1}{1 + z}(I - \Theta X^TK^{-1}X)\beta\\
    &= \frac{1}{1 + z}P_{\Theta; X}^{\perp}\beta.
\end{align}
Therefore
\begin{equation}
    \mathcal{B}(\Theta + \beta\beta^T; \beta, X) = \|P_{\Theta + \beta\beta^T; X}\beta\|_\Sigma^2 = \frac{1}{(1 + z)^2}\|P_{\Theta; X}^{\perp}\beta\|_\Sigma^2 = \frac{1}{(1 + z)^2}\mathcal{B}(\Theta; \beta, X) \le \mathcal{B}(\Theta; \beta, X)
\end{equation}
since $z = u^TK^{-1}u \ge 0$. Therefore $\mathcal{B}(\Theta + \beta\beta^T; \beta) \le \mathcal{B}(\Theta; \beta)$, as desired.
\end{proof}

\subsection{2-D Convolution Derivation} \label{app:2d conv}

In the setting of 2D convolutions, we have $p = s^2$ for $s \times s$ images. First, let us understand the $\mathcal{A}$ operator. For $X \in \mathbb{R}^{n \times n}$, we index the rows by $(i, i')$ (corresponding to the $(i, i')$ pixel in the output channel) and index the columns by $(j, j')$. Thus we can refer to entries in $X$ by $X_{i, i', j, j'}$ for $i, i', j, j' \in [s]$. For a convolutional network with $3 \times 3$ filters, we have $r = 9$ basis elements, which we can index by $(k, k')$ for $k, k' \in [-1, 0, 1]$. In particular, the $(k, k')$ basis element is given by
\begin{equation}
    \{B_{k, k'}\}_{i, i', j', j'} = \delta_{j - i = k, j' - i' = k'}.
\end{equation}
Therefore by a straightforward computation the $\mathcal{A}$ operator is defined as
\begin{equation}
    \mathcal{A}(X)_{i, i', j, j'} = \sum_{(k, k') \in [-1, 0, 1]^2} X_{i + k, i'+k', j + k, j'+k'},
\end{equation}
where again we define $X_{i, i', j, j'} = 0$ if any of $i, i', j, j'$ equal $0$ or $s + 1$. We use this formula to compute the feature transformations used in Figure~\ref{fig: leading directions}.

Observe that $\mathcal{A}$ acts independently on the vector of entries where $(j -i, j'-i')$ is fixed. Let $y^{a, b}$ be the vector of entries where $(j -i, j'-i') = (a, b)$; assuming WLOG that $a, b \ge 0$, define $y^{a, b}_{i, i'} = X_{i, i', i+a, i'+b}$ for $i \in [s - a], j \in [s-b]$. Then, denoting by $A^{a, b}$ the action of $\mathcal{A}$ on $y^{a, b}$, we have that
\begin{equation}
    A^{a, b}y^{a, b}_{i, i'} = \sum_{(k, k') \in [-1, 0, 1]^2} X_{i + k, i' + k'}
\end{equation}
Let $v^a$ be an eigenvector of an $s-a$-dimensional tridiagonal Toeplitz matrix with all nonzero entries being 1, and let $\lambda_a$ be its corresponding eigenvalue. Define the vector $u_{i,i'} = v^a_i v^b_{i'}$. Then:
\begin{align}
    A^{a, b}u_{i, i'} &= \sum_{(k, k') \in [-1, 0, 1]^2} u_{i + k, i' + k'}\\
    &= \sum_{(k, k') \in [-1, 0, 1]^2}v^a_{i + k} v^b_{i' + k'}\\
    &= (\sum_{k \in [-1, 0, 1]}v^a_{i + k})(\sum_{k \in [-1, 0, 1]}v^b_{i' + k})\\
    &= \lambda_a v^a_i \lambda_b v^b_i\\
    &= \lambda_a\lambda_b u_{i, i'}
\end{align}
Therefore $\lambda_a\lambda_b$ is an eigenvalue of $A^{a, b}$. The eigenvalues of $A^{a, b}$ are hence 
\begin{equation}
   \{(1 + 2\cos{\frac{i\pi}{s - a + 1}})(1 + 2\cos{\frac{i'\pi}{s - b + 1}}) : i \in [s-a], i' \in [s-b]\},
\end{equation}
so there is the unique largest eigenvalue $(1 + 2\cos{\frac{\pi}{s - a + 1}})(1 + 2\cos{\frac{\pi}{s - b + 1}})$. The eigenvalues of $\mathcal{A}$ are the eigenvalues of $A^{a, b}$ over all choices of $a, b$, and therefore $\mathcal{A}$ has unique maximum eigenvalue $(1 + 2\cos{\frac{\pi}{s + 1}})^2$. The corresponding eigenvector is when $a = b = 0$, and is thus only supported on the entries $X_{i, i', i, i'}$. Hence the leading eigenvector of $\mathcal{A}$ is diagonal, and simply reweights individual pixels. Using the formula for the eigenvectors of a tridiagonal Toeplitz matrix, the leading eigenvector $\Theta^*$ is given by
\begin{equation}
    \Theta^*_{i, i', i, i'} = \sin\left(\frac{i\pi}{s + 1}\right)\sin\left(\frac{i'\pi}{s + 1}\right)
\end{equation}
It is easy to check that neither of the pooling nor flattening initial conditions are orthogonal to $\Theta^*$, and hence they converge to $\Theta^*$ in the infinite depth limit.

\end{document}